\pdfoutput=1

\documentclass[11pt]{article}

\usepackage[preprint]{acl}
\usepackage{algorithm}
\usepackage{float}
\usepackage{algpseudocode}
\usepackage{amsmath}
\usepackage{enumitem}
\usepackage{xcolor}
\usepackage{amssymb}
\definecolor{Vanilla}{HTML}{ff8d13}
\definecolor{Prompt}{HTML}{bd00ff}
\definecolor{CultureSPA}{HTML}{2782ed}
\definecolor{Modular}{HTML}{545f7f}
\definecolor{Palette}{HTML}{36aa70}
\definecolor{Moe}{HTML}{36aa70}
\definecolor{UserExampleTitle}{HTML}{545f7f}
\usepackage{adjustbox}
\usepackage{times}
\usepackage{latexsym}
\usepackage{epigraph}
\usepackage{array}
\usepackage{booktabs} 
\usepackage{tabularx} 
 \usepackage{latexsym}
\usepackage{tcolorbox}
\usepackage{colortbl}
\usepackage{color}
\usepackage{xcolor}
\usepackage{mdframed}
\usepackage{supertabular}
\newenvironment{fullpagecaption}[1]
{\begin{center}\begin{minipage}{\textwidth}\caption{#1}\end{minipage}\end{center}}
{}
\definecolor{NavyBlue}{rgb}{0.0, 0.0, 0.5}
\definecolor{BlueGreen}{rgb}{0.0, 0.5, 0.5}
\definecolor{Plum}{rgb}{0.5, 0.0, 0.5}
\definecolor{ForestGreen}{rgb}{0.0, 0.26, 0.15}
\definecolor{Red}{rgb}{1.0, 0.0, 0.0}
\usepackage[T1]{fontenc}
\usepackage{soul} 

\definecolor{AfricaColor}{HTML}{66c2a5}
\definecolor{AmericaColor}{HTML}{fc8d62}
\definecolor{AsiaColor}{HTML}{8da0cb}
\definecolor{EuropeColor}{HTML}{e78ac3}
\definecolor{OceaniaColor}{HTML}{a6d854}

\newcommand{\highlightAfrica}[1]{\sethlcolor{AfricaColor}\hl{#1}}
\newcommand{\highlightAmerica}[1]{\sethlcolor{AmericaColor}\hl{#1}}
\newcommand{\highlightAsia}[1]{\sethlcolor{AsiaColor}\hl{#1}}
\newcommand{\highlightEurope}[1]{\sethlcolor{EuropeColor}\hl{#1}}
\newcommand{\highlightOceania}[1]{\sethlcolor{OceaniaColor}\hl{#1}}

\usepackage[utf8]{inputenc}

\usepackage{microtype}

\usepackage{inconsolata}

\usepackage{graphicx}

\title{\includegraphics[width=\baselineskip]{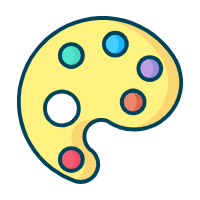}Cultural Palette: Pluralising Culture Alignment via Multi-agent Palette}

\author{
 \textbf{Jiahao Yuan\textsuperscript{1,2,5}},
 \textbf{Zixiang Di\textsuperscript{1,2}},
  \textbf{Shangzixin Zhao\textsuperscript{2,3}},
  \textbf{Zhiqing Cui\textsuperscript{1,4}},
    \textbf{Hanqing Wang\textsuperscript{4}}, \\
   \textbf{Guisong Yang\textsuperscript{2}}, 
 \textbf{Usman Naseem\textsuperscript{5}}
\\
 \textsuperscript{1}ECNU, \quad
 \textsuperscript{2}USST, \quad
    \textsuperscript{3}BNU, \quad
   \textsuperscript{4}HKUST–GZ, \quad
 \textsuperscript{5}MQ
 \\
   \href{mailto:jhyuan.cs@gmail.com}{jhyuan.cs@gmail.com},
   \href{mailto:usman.naseem@mq.edu.au}{usman.naseem@mq.edu.au}
}

\begin{document}
\maketitle
\begin{abstract}
Large language models (LLMs) face challenges in aligning with diverse cultural values despite their remarkable performance in generation, which stems from inherent monocultural biases and difficulties in capturing nuanced cultural semantics. Existing methods struggle to adapt to unknown culture after fine-tuning. Inspired by cultural geography across five continents, we propose \textit{Cultural Palette}, a multi-agent framework that redefines cultural alignment as an adaptive "color-blending" process for country-specific adaptation. Our approach harnesses cultural geography across five continents through three key steps: First, we synthesize the \textit{Pentachromatic Cultural Palette Dataset} using GPT-4o, refining continental-level dialogues with Hofstede's cultural dimensions to establish foundational cultural representations. Second, five continent-level alignment agents form specialized cultural communities that generate region-specific draft responses. Third, a Meta Agent employs \textit{Cultural MoErges} to dynamically blend these cultural "colors" through attention-gated parameter merging, akin to mixing pigments on a palette, resolving conflicts while preserving cultural nuances to produce the final culturally-aligned response. Extensive experiments across various countries demonstrate that \textit{Cultural Palette} surpasses existing baselines in cultural alignment.
\end{abstract}

\section{Introduction}
\begin{figure}[ht]
\centering
\includegraphics[width=0.85\linewidth]{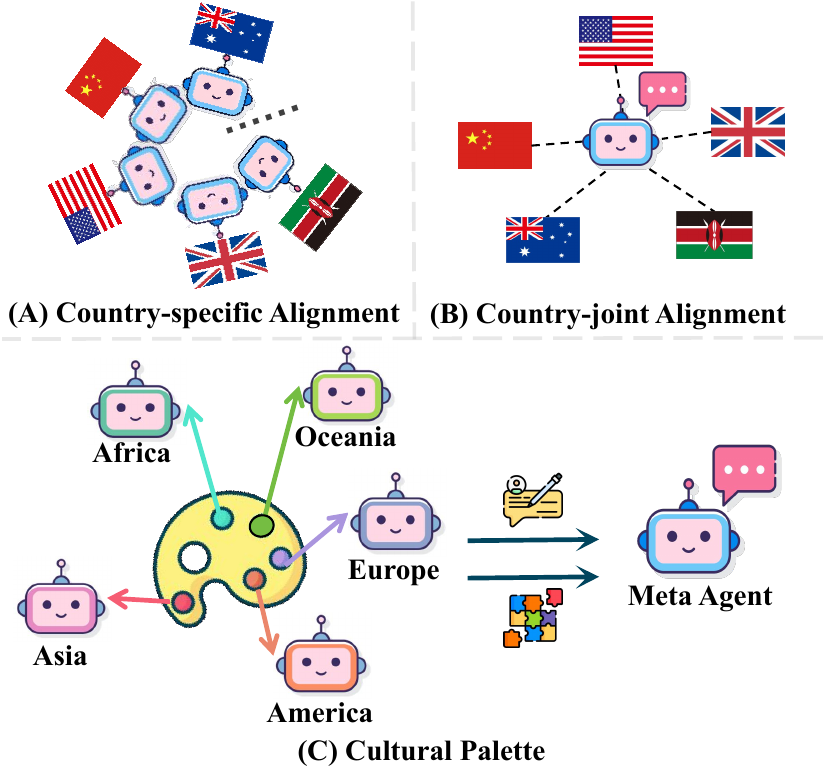}
  \caption{Comparison between (A) country-specific alignment \cite{li2024culturellm,li2024culturepark,feng2024modular}, (B) country-joint alignment \cite{li2024self,li2024culturellm} and (C) Cultural Palette.}
  \label{fig:intro}
  \vspace{-1em}
\end{figure}
Large Language Models (LLMs), such as GPT-4 \cite{achiam2023gpt}, have shown remarkable performance in aligning with human values for real-world applications by incorporating diverse perspectives \cite{han2024value,sorensen2024value} and preferences \cite{ouyang2022training,chen2024pal} through specific fine-tuning and alignment techniques \cite{hu2022lora,ouyang2022training}. However, alignment with human values is predominantly influenced by pluralistic, multicultural factors including traditions, societal norms and more \cite{feng2024modular}, especially in our multicultural society, necessitating LLMs to perceive cross-cultural differences and reflect pluralistic feedback \cite{sorensen2024position,feng2024modular} to avoid mono-cultural bias \cite{shi2024culturebank} risked by mere RLHF \cite{ouyang2022training}. Early studies primarily involves collecting cultural data corpora for fine-tuning to enhance cultural-awareness in LLMs \cite{chan2023harmonizing,lin2024mala,kirk2024prism}, while extremely time-consuming.

Recent research has explored two primary approaches to integrating cultural elements into LLMs: \textbf{(i) multi-culture data composition for alignment}, which simulates social surveys to synthesize culture-specific corpora \cite{li2024culturepark,shi2024culturebank}, cultural value preference choices \cite{li2024culturellm,xu2025self} seeded from the World Values Survey (WVS) \cite{wvs} or PRISM dataset \cite{kirk2024prism} collecting preferences for LLMs responses across cultural regions for fine-tuning and alignment training, which is limited by underrepresented cultures and may lead to overfitting and value leakage from the WVS benchmark \cite{zhou2023don}. \textbf{(ii) multi-culture community mechanisms}, which leverage pluralistic feedback among multi-agent \cite{feng2024modular,lake2024distributional} yet struggles with unseen cultures \cite{feng2024modular}. In summary, pluralistic cultural alignment for LLMs encounters three major limitations: 
\textbf{(1)} Existing alignment techniques like RLHF \cite{ouyang2022training,hong2024orpo} often neglect intersectional identities and marginalized viewpoints between countries instead of among community, limiting the models' ability to provide culturally inclusive outputs.\textbf{(2)} Existing datasets for pluralistic cultural alignment mainly synthesize multiple-choice questions for cultural preference seeded from WVS benchmark for fine-tuning, lacking  cultural dialogues needed for nuanced understanding. 
\textbf{(3)} Cultural norms evolve over time; however, current methods lack effective paradigms to adapt to this fluidity. Requiring fine-tuning on country-specific corpora \cite{li2024culturellm,xu2025self} to construct cultural community \cite{sorensen2024position,feng2024modular} leads to inflexibility and limited practicality \cite{xu2025self}.

To address above limitations, we propose \textit{Cultural Palette}, a framework integrating the core cultural "colors" of five continents \footnote{We adopt the five-continent definition from \citet{buschmann2018world}: Africa, America, Asia, Europe \& Oceania.} \cite{crang2013cultural}, collectively forming a palette that symbolizes global cultural diversity via multi-agent coordination grounded on \textbf{cultural geography findings that culture differences among global countries are profoundly shaped by geographical influences within cultural evolution and spread across five continents }\cite{crang2013cultural}. As depicted in Fig.~\ref{fig:intro}, \textit{Cultural Palette} first generate cultural dialogues based on the \textit{PRISM} dataset \cite{kirk2024prism} for purifying responses representing the cultural characteristics of five continents (Sec.~\ref{sec:dataset}) under Hofstede's cultural dimensions \cite{insights2021hofstede} for perceiving culture differences \cite{hofstede2009geert}, conceptualized as primary colors for cultural alignment (Sec.~\ref{subsec:cultural-orpo}) and then coloring five continent cultures to form a multi-agent cultural community (Sec.~\ref{subsec:multi-agent}). Our key contributions are as follows:
\begin{itemize}
    \setlength{\itemsep}{0pt}
    \item To the best of our knowledge, we introduce the first \textbf{Cultural Palette} framework that models cultural alignment as \textbf{dynamic blending} via \textit{Cultural MoErges}, outperforming four country-specific and joint alignment methods.

    \item We release the \textbf{Pentachromatic Cultural Palette Dataset}, a diverse collection of cultural dialogues synthesized using LLMs and refined using Hofstede dimensions to capture cultural semantics across five continents.
    
    \item We propose a multi-agent framework that \textbf{coordinates five continent-level drafting agents with a meta agent for country-level alignment}. Extensive experiments across 18 countries demonstrate consistent gains over five pluralistic alignment methods.
\end{itemize}
\section{Related Work}
\paragraph{Pluralistic Cultural Alignment.}
Existing alignment techniques, such as DPO \cite{rafailov2024direct} and ORPO \cite{hong2024orpo} guide LLMs to learn from preferred-signal data but do not account for cultural factors. To enhance 
pluralistic culture alignment in LLMs, existing researches focus on aligning with specific countries or values \cite{huang2024acegpt,xu2025self} via finetuning \cite{li2024culturellm,li2024culturepark,huang2024acegpt,xu2025self} and alignment \cite{kirk2024prism,li2024personalized,guo2025care}, yet remains constrained by reliance on inherent cultural values \cite{feng2024modular,vamplew2024multi}. To alleviate the above dependencies, recent research leverages LLMs' zero-shot capabilities \cite{kojima2022large,li2024culturepark} for cultural alignment through multi-agent collaboration, introducing a cultural-aligned judger LLM to summarize community comments via Overton, steerable, and distributional modes \cite{sorensen2024position,feng2024modular,shetty-etal-2025-vital}, while challenges persist in interacting with unseen culture \cite{feng2024modular}. To further address the complexities of cultural diversity, we propose a novel approach inspired by cultural evolution across five continents \cite{crang2013cultural}. By conceptualizing these continental cultures as fundamental colors on a palette, we introduce a multi-agent framework integrated with a \textit{Cultural MoErges }\cite{goddard2024arcee} mechanism integrating semantic relationships at both continent and country levels to dynamically blend colors adapting to judging diverse cultural alignment and ensuring nuanced and context-aware response through multi-agent collaboration.

\paragraph{Data Synthesis for Cultural Alignment.}
Aligning LLMs for cultural pluralism—whether through fine-tuning \cite{, li2024culturellm, li2024culturepark, shi2024culturebank}, alignment \cite{kirk2024prism, li2024personalized,guo2025care}, or agent-based approaches \cite{sorensen2024position, feng2024modular}—requires extensive, culture-specific datasets. While existing datasets like PRISM \cite{kirk2024prism} collect user feedback on LLM responses across diverse countries, they primarily focus on preference ranking rather than generating culturally nuanced dialogues, limiting their utility for training models to understand implicit cultural norms \cite{chen2024pal}. Recent studies have demonstrated that self-instruct \cite{wang2023self} or multi-agent \cite{li2024culturellm} can effectively synthesize culturally nuanced data through LLM-driven multi-step generation and refinement, including expanding datasets based on the World Values Survey (WVS) \cite{wvs} to study cultural dominance and alignment, with benchmarks such as CultureLLM \cite{li2024culturellm}, CulturePark \cite{li2024culturepark}, and CultureSPA \cite{xu2025self}. However, LLM-driven data synthesis, seeded from the WVS for multiple-choice data pairs, may introduce biases in cultural options and lead to value leakage from the WVS benchmark \cite{zheng2023large, zhou2023don}. To overcome these limitations, we expanded Prism \cite{kirk2024prism} to cultural dialogues from five continents, creating \textit{Pentachromatic Cultural Palette Dataset} (Sec.~\ref{sec:dataset}) through self-feedback contrastive aggregation of cultural differences. 

\paragraph{Model Merging.}
LLMs trained on a single corpus often suffer from limited generalization and suboptimal decision-making, as they may converge to local optima that restrict efficiency and versatility across tasks \cite{lawson2024merging,yang2024model}. To address this, prior work explores collaborative model merging strategies—such as Task Arithmetic \cite{ilharcoediting}, Ties \cite{yadav2024ties}, and Model Stock \cite{jang2025model}—to improve performance on complex decision-making \cite{lawson2024merging} and generation tasks \cite{pourreza2024sql}. More recently, MoErges \cite{goddard2024arcee,tang2024merging,jiang2024mixtral} unifies model merging with Mixture-of-Experts architectures through an untrainable design, initializing MoE MLP layers with task-specific model weights. Building on these advances, we investigate the role of model merging in culturally grounded decision-making by systematically comparing merging-based methods with cultural joint training approaches \cite{xu2025self,li2024culturellm}. Furthermore, we ground our proposed \textit{Cultural MoErges} in palette theory, demonstrating how continent-level cultural representations can be dynamically composed to foster country-level cultural pluralism.
\begin{figure*}[ht]
  \centering
  \includegraphics[width=0.95\linewidth]{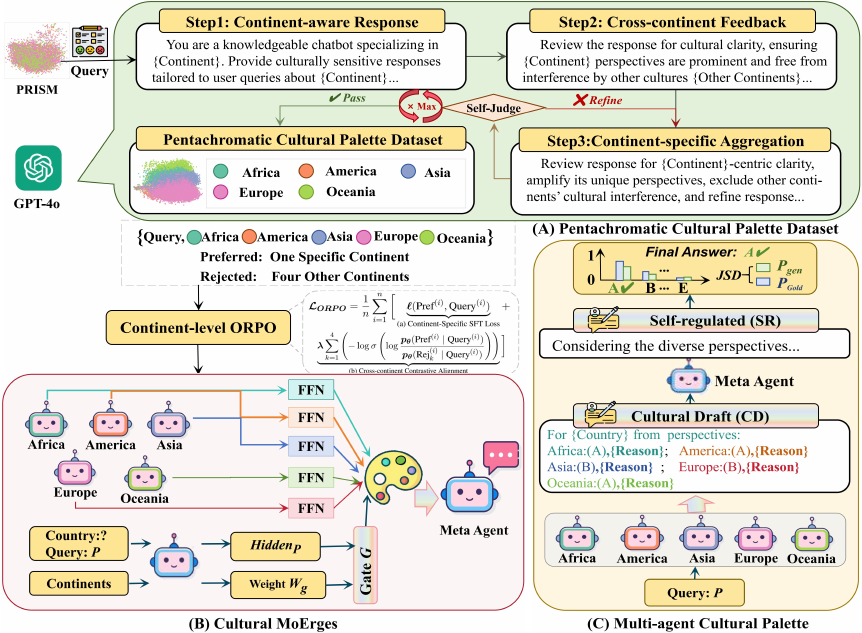}
  \caption{Architecture of Cultural Palette consisting of continent-level alignment on Pentachromatic Cultural Palette Dataset (Sec.~\ref{sec:dataset}), Cultural MoErges for Meta Agent (Sec.~\ref{subsec:cultural-orpo}) and Multi-agent Cultural Palette (Sec.~\ref{subsec:multi-agent}). Details of our ORPO alignment loss formulation are deferred to Appendix B for brevity.}
  \label{fig:model}
\end{figure*}
\section{Pentachromatic Cultural Palette Dataset}
\label{sec:dataset}
We contribute a Pentachromatic Cultural Palette dataset with the diverse cultural semantics and values across five continents in social dialogues. 
\paragraph{Dataset Construction.}
Cultural differences in evolution across five continents \cite{ crang2013cultural} shape values and semantics in different countries. While existing datasets like PRISM \cite{kirk2024prism} collect preference rankings, they lack nuanced dialogues necessary for training models to understand context-specific norms. To address this, we propose a Pentachromatic Cultural Palette dataset, built from PRISM, to capture cultural nuances essential for reflecting the full diversity of cultural semantics in social dialogues. Following \cite{li2024self}, We enhance cultural richness through a multi-step data synthesis process powered by \texttt{GPT-4o}, utilizing user queries $\{Query_{j}\}_{j=1}^{n}$ from the PRISM dataset. As shown in Fig.~\ref{fig:model}, our pipeline begins with a self-instruct \cite{wang2023self} step to generate initial continent-aware responses. These responses then undergo a critique phase based on cross-continent feedback to identify divergent norms. Finally, in the continent-specific aggregation stage, each response is polished through an iterative self-refinement loop \cite{madaan2024self}. Within this loop, the model acts as a self-judge to iteratively enhance the response, ensuring cultural relevance while preventing cross-cultural contradictions. This process is capped at a maximum of $Max=3$ rounds and terminates early once a response passes predefined quality checks. And finally we get the Pentachromatic Cultural Palette Dataset \footnote{Appendix~\ref{subsec:dataprompt} details the multi-step synthesis prompts, Appendix~\ref{subsec:distribution} reports dataset scale, and Fig.~\ref{fig:dataset} validates the five-continent base colors. All synthesized data undergoes rigorous human ethics review for cultural fairness and safety \cite{kirk2024prism,li2024culturellm}.},  represented as $\{(Query_{j}, Response_{j}^{continent})\}_{j=1}^{n}$, where $continent \in$ \{\textit{Africa}, \textit{Asia}, 
 \textit{Europe},  \textit{America}, \textit{Oceania}\}. 

\section{The Cultural Palette Framework}
\label{sec:palette}
Following \citet{wang2025mixture}, our multi-agent cultural palette encompasses a country-specific Meta Agent (Sec.~\ref{subsec:cultural-orpo}) aggregating drafts from five continent-level cultural communities (\ref{subsec:multi-agent}). 





\subsection{Cultural Palette for Meta Agent}
\label{subsec:cultural-orpo}
\paragraph{Joint/Merging Alignment.} After constructing the Pentachromatic Cultural Palette dataset, we leverage it to perform continent-level alignment for LLMs as cultural decision-maker judger aligned with users, ensuring culturally relevant outputs tailored to diverse regions. Inspired by \citet{li2024self}, we introduce two alignment strategies to highlight the benefits of model merging for overall cultural pluralism. The first strategy, joint alignment, integrates all cultural dialogue data into a single LLM. The second strategy involves creating separate models for each culture, which are then merged into a unified LLM using separately various merging techniques including Ties \cite{yadav2024ties}, Model Stock \cite{jang2025model} and MoErges \cite{goddard2024arcee,jiang2024mixtral} to demonstrate that MoErges exhibits the greatest potential for cultural alignment (see Sec.~\ref{subsec:align}). 

\paragraph{Cultural MoErges (CM) for Meta Agent.} Inspired by the merging paradigm of MoErges \cite{goddard2024arcee,tang2024merging}, we propose Cultural MoErges for Meta Agent to merge cultural knowledge of all five continent-level experts as a shared resource, enabling dynamic adaptation to country-specific contexts following \citet{goddard2024arcee} while acting as a global cultural reviewer. Five continent-aligned models $\mathcal{E}_{\text{Group}} = \{E_{\text{Africa}}, E_{\text{Asia}}, E_{\text{Europe}}, E_{\text{America}}, E_{\text{Oceania}}\}$ are trained on region-specific datasets, sharing a base LLM architecture but differing in FFN components following \cite{feng2024modular}, and the activation mechanism dynamically adjusts expert contributions through a continent-to-country aware gate that computes weights $G(P) \in \mathbb{R}^5$ to prioritize expert contributions based on national cultural cues in user prompt $P$ via: $G(P) = \text{Softmax}(hidden_P \cdot W_g)$, with $W_g \in \mathbb{R}^{d \times 5}$ initialized via hidden states of continent-level system prompts (e.g., "You are a knowledge chatbot about \{continent\}...") and $hidden_P$ derived from the base LLM's encoding of $P$ such that $G(P)$ blend experts' parameters into $LLM_{\text{meta}}$'s architecture following MoErges paradigm \cite{wang2025mixture}, the Meta Agent's parameter space $\Theta_{\text{meta}}$ is defined as:
\begin{equation}
\Theta_{\text{meta}} = \Theta_{\text{shared}} \oplus \left( \sum_{C \in \mathcal{E}_{\text{Group}}} G(P) \cdot \Theta_C^{\text{FFN}} \right).
\label{eq:fusion}
\end{equation}
where $\Theta_{\text{shared}} \subset \Theta_{\text{base}}$ denotes frozen shared parameters inherited from the base LLM and $\Theta_C^{\text{FFN}}$ represents the FFN parameters of continent-level expert $C$, dynamically fused via $\text{Softmax}$-normalized weights $G(P)$. Empirical experiments in Sec.~\ref{subsec:mainresult} demonstrate that Cultural MoErges outperforms alternative strategies in achieving both cultural pluralism and country-specific alignment.

\subsection{Multi-agent Cultural Palette}
\label{subsec:multi-agent}
As illustrated in Fig.~\ref{fig:model}, multi-agent cultural palette consist of draft from community comments aligned with five continents, self-regulated aggregation and final decision from Meta Agent. 
\paragraph{Stage 1: Cultural Draft (CD).}

Given a country-specific question $P$ that includes a target country $C$ and a query $Q$, we engage a continent-level LMs community $\mathcal{E}_{Group} = \{E_{Africa}, E_{Asia}, E_{Europe}, E_{America}, E_{Oceania}\}$, each acted as a proxy for a regional community voice informed by local norms, rhetorical patterns, and sociocultural context. While $C$ is often closely affiliated with one or more continents, all community LMs are jointly activated to span both proximate and distal cultural standpoints aligned with pluralistic roadmap~\cite{sorensen2024position}, promoting culturally diverse yet contextually coherent perspectives \cite{crang2013cultural}, forming a pluralistic Cultural Draft $Draft$ via culture-aware prompting $P_{draft}$:

\begin{small}
\begin{equation}
Draft = \sum_{E \in \mathcal{E}{\text{Group}}} Continent \oplus E(Q, P_{draft}).
\end{equation}
\end{small}
where $\oplus$ denotes the association of the response with its continent tag $Continent$.

\paragraph{Stage 2: Self-regulated Aggregation (SR).} 
While continent-level LMs in the community contribute diverse—and occasionally conflicting—cultural standpoints, the objective is not to reconcile these differences but to reinterpret them through the lens of the target country, thereby supporting downstream alignment with the country-specific response distribution. To this end, we introduce a \textit{Self-regulated Aggregation} mechanism, wherein $LLM_{\text{meta}}$, activated under country $C$ via Cultural Merge parameter space (Sec.~\ref{subsec:cultural-orpo}), recontextualizes each regional draft according to culturally encoded priors aligned with $C$. Rather than flattening disagreement, $LLM_{\text{meta}}$ preserves both consonant and dissonant perspectives, reshaping them through a culturally grounded filter that centers $C$'s normative worldview. Inspired by reflective adaptation principles in formative learning~\cite{nicol2006formative}, $LLM_{\text{meta}}$ serves as a cultural interpreter—transforming heterogeneous inputs into a coherent intermediate response, denoted as $Answer_{\text{regulate}}$, that preserves cross-cultural plurality while aligning with the socio-contextual frame of country $C$.
\paragraph{Stage 3: Final Decision.}
Conditioned on the intermediate output $Answer_{\text{regulate}}$, the meta agent $LLM_{\text{meta}}$ synthesizes a culturally grounded response using the final decision prompt $P_{\text{final}}$. Leveraging the aggregated drafts as contextual input, it generates a country-specific answer $Answer_{final}$ and distribution $P_{\text{gen}}$ that reflects the normative worldview of target country $C$, supporting post-hoc cultural alignment evaluation by comparing $P_{\text{gen}}$ with empirical response distributions $P_{\text{gold}}$ (Sec.~\ref{subsec:align}), ultimately enabling continent-to-country cultural alignment.

\section{Experiments}

\subsection{Models and Implementation}
\begin{figure}[t]
  \centering
\includegraphics[width=0.86\linewidth]{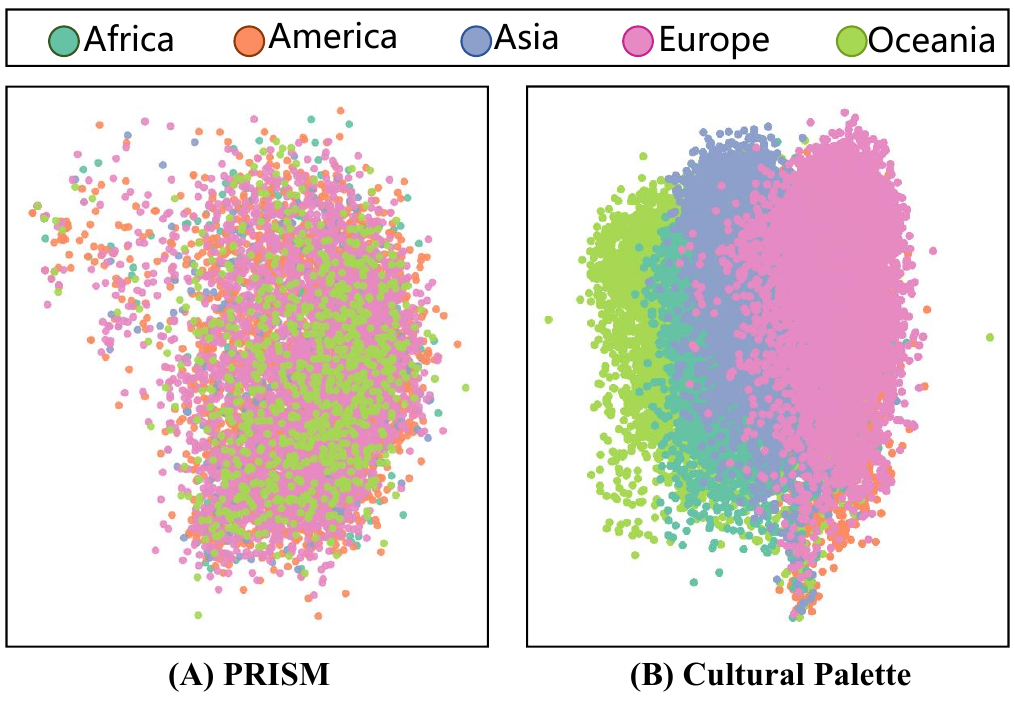}
\caption{Comparative analysis of semantic representations on the PRISM and Pentachromatic Cultural Palette datasets using Multilingual embeddings (\texttt{microsoft/Multilingual-MiniLM-L12-H384}) with PCA-based dimensionality reduction. Data points are color-coded by continent, based on the country associated with each QA pair.}
  \label{fig:dataset}
  \vspace{-1.2em}
\end{figure}
\paragraph{Data  Synthesis.}We utilized the original questions from the PRISM dataset \cite{kirk2024prism}, which comprises 7,805 questions sourced from 75 birth countries. To achieve balanced representation, we augmented it by synthesizing 7,805 questions per continent, ensuring equal distribution among five continents and resulting in a total of 39,025 question-answer pairs, with each continent having 31220 preference pairs derived from the interactions between every two continents. As shown in Fig.~\ref{fig:dataset}, the PRISM dataset’s substantial distributional overlap across continents underscores its limitations in capturing cultural distinctions, posing challenges for modeling nuanced cultural segmentation and aligning with region-specific applications. In contrast, our Pentachromatic Cultural Palette dataset demonstrates superior clustering, with data points forming distinct, continent-specific groups that effectively model cultural uniqueness and enable more accurate, culturally aligned representations. The distribution of topics and continents are detailed in Appendix A.2.

\paragraph{Cultural Alignment.} We conducted all experiments using Llama3.1-8B-Instruct \cite{dubey2024llama} and Qwen2.5-7b-Instruct \cite{yang2024qwen2}. Additionally, we employed five continent-specific alignments with ORPO \cite{hong2024orpo}, a parameter-efficient monolithic alignment method, on our Pentachromatic Cultural Palette Dataset, utilizing MS-Swift \cite{zhao2025swift} for alignment and MergeKit \cite{goddard2024arcee} for merging on two \texttt{NVIDIA A100} GPUs (see Appendix B for experiment setting \& alignment loss details). For model training and evaluation, we partitioned our dataset was split 9:1 for training and testing, with the test set evaluating Cultural MoErges against alternative alignment methods.
\begin{figure*}[ht]
  \centering
  \includegraphics[width=0.95\linewidth]{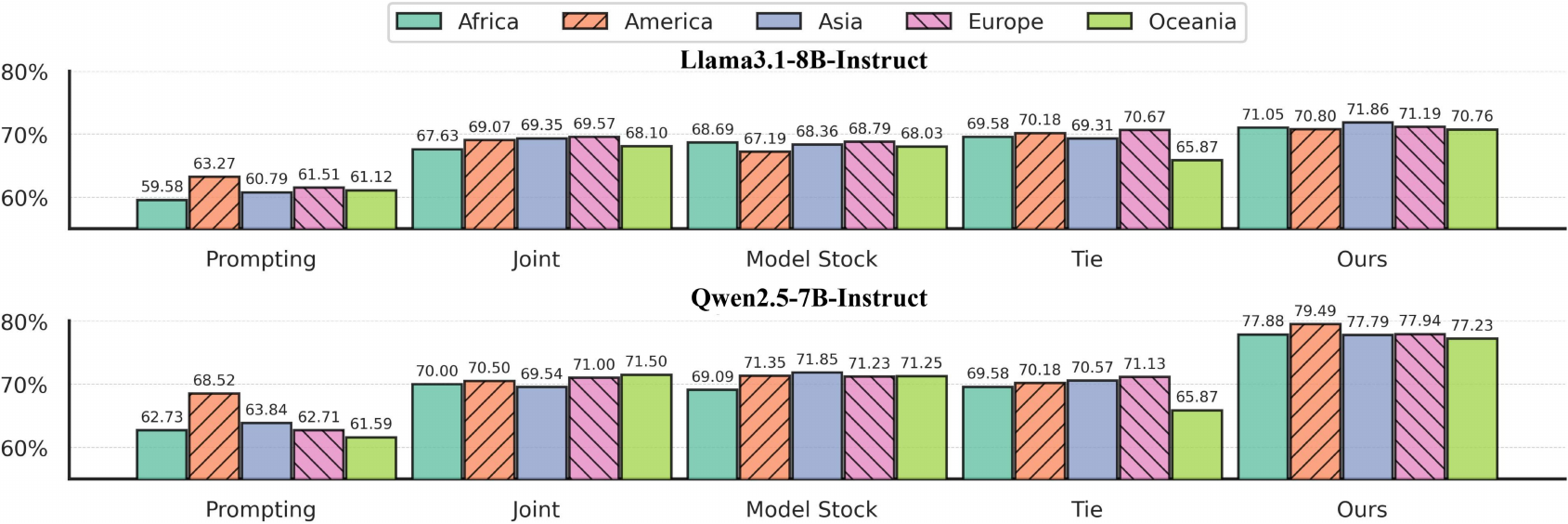}
\caption{Comparison of semantic-consistency alignment score between our  Cultural Palette and other methods (Prompting, Joint) and merging strategies (Tie and Model Stock) on Llama3.1-8B-Instruct and Qwen2.5-7b-Instruct.}
  \label{fig:align}
\end{figure*}
\subsection{Datasets and Tasks}
\label{subsec:align}

\paragraph{Task 1: Cultural Alignment \textit{w/ NLI evaluation.}} To evaluate Cultural MoErges against alternative merging methods and alignment strategies, we measure LLMs’ ability to generate continent-level responses aligned with cultural values. Following \citet{feng2024modular}, we use a DeBERTa-v3-large NLI model fine-tuned on standard NLI datasets \cite{sileo-2024-tasksource} to compute a semantic-consistency score as the Entailment probability between gold values $R_{gold}$ and LLM-generated responses $R_{llm}$:

\begin{small}
\begin{equation}
   S_{semantic} = P(\text{Entailment} \mid R_{gold}, R_{llm}) \times  100 \%  \in [0,100],
\end{equation}
\end{small}

where higher percentages indicate stronger alignment with the gold-standard cultural values.

\paragraph{Task 2: Cultural Palette \textit{w/ GlobalOpinionQA.}} Following \citet{feng2024modular,xu2025self}, we compute alignment scores $S_{align}$ for 18 countries shown in Table~\ref{tab:align_res} across five continents using the GlobalOpinionQA dataset \cite{durmus2024towards}, which aggregates global survey data. For each country-level survey question, LLMs are prompted to integrate national context and generate probability distributions $P_{gen}$ over answer options, which are then compared to ground-truth survey response distribution $P_{gold}$ via Jensen-Shannon distance to quantify cultural alignment:

\begin{small}
\begin{equation}
    S_{align} = 1 - \frac{1}{2} \texttt{KL}(P_{gen} \parallel M) - \frac{1}{2} \texttt{KL}(P_{gold} \parallel M).
\end{equation}
\end{small}

where $M$ is the average distribution of $P_{gen}$ and $P_{gold}$, and \texttt{KL} denotes the Kullback-Leibler divergence. Higher $S_{align}$ indicates better cultural alignment with  real-world survey distributions.

\subsection{Baselines}
We first evaluated Cultural MoErges against four baseline alignment strategies: \textbf{(1) Prompting}: We promote cultural alignment by appending instructions via the same prompting template as culture-aware prompting following \cite{feng2024modular,xu2025self}; \textbf{(2) Joint}: Following \citet{xu2025self}, we implement culturally joint alignment with ORPO \cite{hong2024orpo} for overall cultural LLM alignment; 
\textbf{(3) Tie}: Following \citet{yadav2024ties}, we merge LLMs among continent-specific models by resetting minor fine-tuned parameters, resolving sign conflicts, and merging parameters with consistent signs; \textbf{(4) Model Stock}: effectively integrate fine-tuned LLMs across continent-specific applications \cite{jang2025model}. 

Subsequently, we compare Cultural Palette with five classical and latest baselines on Llama3.1-8B-Instruct \cite{dubey2024llama} and Qwen2.5-7b-Instruct \cite{yang2024qwen2}: \textbf{(1) Vanilla}: Following \cite{feng2024modular}, we directly employ LLM to generate responses according to user queries without any modifications. \textbf{(2) Prompting}: We follow \cite{feng2024modular} in promoting cultural pluralism with the same instruction “Make sure your response reflects diverse values and perspectives”. \textbf{(3) Joint}: \citet{xu2025self} adopted joint cultural alignment for comprehensive LLM alignment with culture-aware prompting.  \textbf{(4) Mixture-of-Experts (MoE)}: User queries are matched to the best-suited community LM, which generates culturally aligned responses.
\textbf{(5) Modular}: \citet{feng2024modular} proposed three forms of multi-agent collaboration, including Overton, steerable, and distributional modes.
\subsection{Main Result}
\label{subsec:mainresult}
\begin{table*}[!htbp]
\centering
\renewcommand{\arraystretch}{1.2} 
\begin{adjustbox}{max width=1\textwidth}
\large
\begin{tabular}{c|cccc|cccc|cccc|cccc|cc|c}
\toprule
{} & \multicolumn{4}{c|}{\textbf{America}} & \multicolumn{4}{c|}{\textbf{Europe}} & \multicolumn{4}{c|}{\textbf{Asia}} & \multicolumn{4}{c|}{\textbf{Africa}} & \multicolumn{2}{c|}{\textbf{Oceania}} & \textbf{Avg} \\
\cmidrule(lr){2-19}
{} & \textbf{USA} & \textbf{CAN} & \textbf{BOL} & \textbf{BRA} & \textbf{GBR} & \textbf{NLD} & \textbf{DEU} & \textbf{UKR} & \textbf{CHN} & \textbf{RUS} & \textbf{IND} & \textbf{THA} & \textbf{KEN} & \textbf{NGA} & \textbf{ETH} & \textbf{ZWE} & \textbf{AUS} & \textbf{NZL} & {} \\
\midrule
{\textbf{GPT-4o}} &72.47 &70.28 &72.63 &69.44 &71.01 &69.09 &73.04 &72.96 &72.18 &74.45 &72.16 &72.12 &73.50 &72.16 &72.39 &74.40 &70.44 &69.08 &71.88 \\
\midrule
\multicolumn{19}{c}{\textsc{\textbf{Llama3.1-8B-Instruct}} \cite{dubey2024llama}} \\
\midrule
\textcolor{Vanilla}{\textbf{Vanilla}} &67.81 &65.46 &66.46 &66.88 &59.21 &64.30 &63.95 &59.13 &62.76 &62.72 &66.18 &63.01 &67.54 &66.80
&67.31 &58.22 &64.17 &55.68 &63.76 \\
\textcolor{Prompt}{\textbf{Prompting}} &69.18 & 67.81 & 69.43 & 68.64 & 60.90 & 66.65 & 67.46 & 63.15 & 65.81 & 68.09 & 69.67 & 66.22 & 69.12 & 68.50 & 69.51 & 59.48 & 67.59 & 58.49 & 66.43 \\
\textcolor{CultureSPA}{\textbf{Joint}} &70.10 & 69.24 & 70.94 & 66.35 & 66.93 & 70.27 & 70.54 & 66.11 & 67.41 & 69.21 & 70.12 & 63.29 & 67.14 & 68.44 & 66.75 & 61.34 & 68.41 & 60.06 & 67.37 \\
\textcolor{Moe}{Moe} &73.91 & 70.97 & 76.09 & 75.50 & 73.92 & \underline{76.69} & 72.48 & 72.87 & 76.52 & 74.41 & 73.02 & \underline{75.07} & 74.85 & \textbf{78.76} & 69.57 & 70.30 & 65.25 & 62.77 & 72.94 \\
\textcolor{Modular}{\textbf{Modular}}
&75.14 & 75.70 & 76.60 & \textbf{76.29} & \underline{76.46} & \textbf{76.71} & 74.98 & 71.69 & 74.16 & 73.56 & \underline{74.21} & 71.76 & 71.63 & 70.03 & 70.05 & \underline{74.80} & \underline{72.42} & \underline{72.57} & 73.82 \\
\rowcolor[RGB]{236,244,252}
\;\; \textbf{Ours} &\textbf{75.85} & \textbf{77.56} & \textbf{77.88} & \underline{75.83} & \textbf{76.48} & 76.53 & \underline{76.38} & \textbf{77.69} & \underline{77.12} &\textbf{ 77.67} & \textbf{74.32} & \textbf{75.54} & 76.12 & \underline{77.07} & \textbf{73.65} & \textbf{74.82} & 72.14 & \textbf{73.50} & \textbf{75.90}
 \\
 \midrule
 \textbf{w /o \textit{CM}} &70.12 & 72.42 & 72.77 & 72.37 & 69.31 & 71.01 & 71.05 & 70.12 & 71.52 & 72.39 & 71.84 & 69.36 & 70.68 & 70.57 & 66.23 & 69.66 & 69.80 & 69.51 &70.60 \\
 \textbf{w /o \textit{CD}} &72.25 & 73.77 & 73.80 & 74.27 & 62.23 & 74.47 & 73.25 & 72.37 & 73.70 & 75.97 & 75.99 & 71.90 & \textbf{76.19} & 72.94 & 67.74 & 60.09 & \textbf{73.44} & 69.01 &71.85
\\
 \textbf{w /o \textit{SR}} &\underline{75.54} & \underline{77.03} & \underline{77.39} & 75.25 & 75.27 & 76.32 & \textbf{76.54} & \underline{77.19} & \textbf{77.87} & \underline{77.42} & 73.49 & 74.57 & 74.73 & 76.08 & \underline{71.93} & 73.89 & 71.67 & 71.04 & \underline{75.18}
\\
\midrule
\multicolumn{19}{c}{\textsc{\textbf{Qwen2.5-7B-Instruct}} \cite{yang2024qwen2}} \\
\midrule
\textcolor{Vanilla}{\textbf{Vanilla}} &61.50 & 60.54 & 62.67 & 63.22 & 63.86 & 63.66 & 61.05 & 59.32 & 63.78 & 60.77 & 60.62 & 62.12 & 59.87 & 63.94 & 63.56 & 57.43 & 54.40 & 58.84 & 61.18
 \\
\textcolor{Prompt}{\textbf{Prompting}} &63.18 & 63.80 & 63.48 & 64.76 & 64.77 & 64.52 & 61.86 & 62.75 & 64.99 & 61.42 & 61.35 & 62.20 & 61.60 & 63.15 & 61.39 & 56.77 & 52.05 & 54.44 & 61.58
 \\
\textcolor{CultureSPA}{\textbf{Joint}} & 70.00 & 67.47 & 65.03 & 67.36 & 66.90 & 68.74 & 68.22 
& 63.21  & 63.27 & 62.80 & 69.25 & 65.77 & 69.36  & 69.35  & 66.67 
& 66.20  & 68.00 & 62.79  & 66.69 \\ 
\textcolor{Moe}{\textbf{Moe}} &\underline{73.56} & \underline{72.59} & 74.31 & 71.07 & \textbf{70.96} & 68.47 & 68.93 & 66.76 & 68.67 & 68.14 & 66.15 & 68.65 & 66.17 & 69.51 & \underline{70.63} & 65.99 & \underline{70.51} & 65.49 & 69.25 \\
\textcolor{Modular}{\textbf{Modular}} &72.64 & 68.03 & 72.43 & 73.66 & 69.21 & \textbf{71.21} & 72.78 & \underline{75.57} & 73.01 & 74.79 & 66.62 & \underline{72.19} & 69.44 & 70.32 & 62.04 & \underline{70.52} & 57.58 & 59.09 & 69.51
 \\
\rowcolor[RGB]{236,244,252}
\;\; \textbf{Ours} &\textbf{75.50} & \textbf{73.92} & \textbf{78.83} & \textbf{76.71} & \underline{69.77} & \underline{70.89} & \textbf{73.91} & \textbf{75.58} & \underline{73.46} & \textbf{76.09} & \underline{72.87} & \textbf{73.02} & \textbf{76.51} & \textbf{74.51} & 70.30 & 69.25 & \textbf{74.85} & \textbf{69.57}
&\textbf{73.64}
\\
 \midrule
 \textbf{w /o \textit{CM}} &68.79 & 67.46 & 69.63 & 68.93 & 62.97 & 66.19 & 69.48 & 68.56 & 68.83 & 68.01 & 69.26 & 66.57 & 67.45 & 66.77 & 62.74 & 65.47 & 66.87 & 67.16 &67.29 
\\
 \textbf{w /o \textit{CD}} &72.62 & 68.22 & 68.52 & 71.35 & 49.30 & 64.55 & 71.96 & 72.77 & 71.79 & \underline{74.86} & 73.98 & 63.07 & 69.95 & 72.50 & 62.86 & 46.48 & 68.05 & 58.67 & 66.75
\\ 
\textbf{w /o \textit{SR}} &72.45 & 70.84 & 72.78 & 72.62 & 67.36 & 69.24 & 71.83 & 72.05 & \textbf{73.68} & 74.26 & \textbf{73.06} & 70.25 & \underline{72.11} & \underline{73.30} & \textbf{72.53} & \textbf{74.24} & 69.85& \underline{68.05} & \underline{71.69}
\\
\bottomrule
\end{tabular}
\end{adjustbox}
\caption{Comparison of culture alignment score (converted to percentages) between Cultural Palette and baselines, evaluated across 18 countries grouped by region: (1) America: USA (United States), CAN (Canada), BOL (Bolivia), BRA (Brazil); (2) Europe: GBR (United Kingdom), NLD (Netherlands), DEU (Germany), UKR (Ukraine); (3) Asia: CHN (China), RUS (Russia), IND (India), THA (Thailand); (4) Africa: KEN (Kenya), NGA (Nigeria), ETH (Ethiopia), ZWE (Zimbabwe); (5) Oceania: AUS (Australia), NZL (New Zealand).}
\label{tab:align_res}
\end{table*}
\paragraph{Cultural MoErges achieves superior alignment with continent-level values and perspectives.}
As shown in Fig.~\ref{fig:align}, Cultural MoErges consistently achieves the strongest overall semantic-alignment performance across continents, outperforming both alignment-based and parameter-merging baselines. Prompting and Joint alignment methods exhibit limited regional adaptability, yielding substantially lower scores under continent-specific evaluation (e.g., 59.58\% in Africa and 60.79\% in Asia under Prompting on LLaMA-based models). Parameter-merging approaches, including Model Stock and Tie, provide moderate improvements but display noticeable cross-continent inconsistency (e.g., 71.13\% in Europe versus 65.87\% in Oceania under Tie on Qwen2.5-7B-Instruct). In contrast, Cultural MoErges enables coherent integration of continent-level reasoning, resulting in consistent gains across regions. Specifically, it achieves 71.05\% in Africa and 70.76\% in Oceania on LLaMA-based models, and 77.88\%–79.49\% on Qwen2.5-7B-Instruct, surpassing strong modular and merging baselines by approximately 1.5–3.0 percentage points. These results indicate that Cultural MoErges effectively captures continent-level cultural structure while maintaining semantic consistency. A qualitative case study for culture alignment on question-answering \textit{w/ NLI evaluation} is detailed in Appendix~\ref{subsec:nli}.

\paragraph{Cultural Palette achieves superior real-world cultural alignment.}
\textbf{(1) Best Performance:} As reported in Table~\ref{tab:align_res}, Cultural Palette consistently attains the highest cultural alignment scores across two LLMs, outperforming all baselines in 18 countries spanning five continents. On LLaMA3.1-8B-Instruct, Cultural Palette achieves an average score of 75.90\%, surpassing the strongest baseline Modular (73.82\%) by 2.08 percentage points, with pronounced gains in Asia (3.73) and Africa (5.28), regions where prior methods often struggle to capture cultural nuance. On Qwen2.5-7B-Instruct, the improvement is even more significant: Cultural Palette reaches 73.64\%, exceeding Modular by 4.13 percentage points, with consistent gains across America (6.35), Europe (0.48), and Asia (3.08). \textbf{(2) Alignment Fidelity:} Beyond $S_{align}$ \cite{feng2024modular,xu2025self}, we evaluate distributional fidelity using \textbf{Pearson correlation} with ground-truth survey data. Modular, which relies on country-specific supervision, generalizes poorly (Pearson $r=-0.044$). In contrast, Cultural Palette, trained with continent-level priors and transferred to individual countries, achieves substantially higher correlation (r=0.555). This demonstrates that coarse-to-fine regional transfer improves stability and generalization while outperforming direct country-level learning, which is prone to data sparsity and overfitting. Detailed case studies on GlobalOpinionQA and correlation analyses are provided in Appendix~\ref{subsec:nli} and Appendix~\ref{subsec:pearson}. 
\textbf{(3) GPT-4o as Cultural Teacher:} GPT-4o attains a baseline alignment score of 71.88\%, and our synthetic dataset (Sec.~\ref{sec:dataset}) effectively enhances SLMs’ cultural alignment, surpassing GPT-4o itself in several regions.

\paragraph{Ablation Study.}
To quantify the contribution of each component, we perform three ablations (Table~\ref{tab:align_res}): (i) \textbf{w/o \textit{Cultural MoErges (CM)}}: removing cross-cultural adaptation via CM; (ii) \textbf{w/o \textit{Cultural Draft (CD)}}: excluding continent-level cultural explanations; and (iii) \textbf{w/o \textit{Self-Regulate (SR)}}: eliminating the self-regulation mechanism that integrates multiple drafts. Removing \textbf{CM} causes the largest performance drop: on Qwen2.5-7B-Instruct, the average score decreases from 73.64\% to 67.29\%, with substantial declines in Asia (4.03 percentage points) and Oceania (4.40 percentage points). This highlights CM’s key role in grounding country-level alignment through continent-level priors, which provide semantic regularization and enable cross-national transfer under data scarcity or noisy cultural signals. Ablating \textbf{CD} also results in notable losses, particularly in semantically diverse regions such as the Americas (3.70) and Europe (6.91), indicating that CD expands the model’s hypothesis space with grounded alternatives that guide downstream reasoning. Removing \textbf{SR} produces smaller but consistent drops, with an average decrease of 1.95 percentage points, emphasizing its role in integrating multiple drafts into a coherent cultural response. These findings validate our multi-agent alignment strategy: \textbf{CD} enriches semantic priors with diverse perspectives, \textbf{CM} anchors them through coarse-to-fine structural transfer, and \textbf{SR} fuses divergent outputs into a culturally coherent response, together enhancing and stabilizing culture alignment.

\section{Conclusion}
In this work, we propose Cultural Palette, a novel multi-agent framework for culturally pluralistic alignment in LLMs, modeling five continental cultures as primary colors. Powered by the Pentachromatic Cultural Palette Dataset and a self-regulating \textit{Cultural MoErges} mechanism, Cultural Palette dynamically aggregates continent-level agents to adapt across cultural contexts. Empirical results reveal three key findings: (1) Cultural MoErges enables dynamic composition of continent-level agents, surpassing four modular and training-based baselines in semantic consistency and cultural adaptability; (2) Synthesizing diverse continental data alone suffices to achieve fine-grained continent-to-country cultural alignment, enabling practical adaptation to country-level nuances without the need for country-specific data collection; and (3) Cultural Palette consistently outperforms five pluralistic alignment baselines across both semantic and cultural alignment metrics. Our research may inspire future studies on modeling pluralistic alignment by leveraging continent-level agents to adapt to specific countries and their unique cultural contexts.
\section*{Limitations}
Our Cultural Palette integrates the core cultural "colors" of five continents with a Meta Agent that blends these base colors into a unified multi-agent framework, symbolizing global cultural diversity \cite{crang2013cultural}. Extensive experiments and case studies show that our multi-agent framework enhances the cultural sensitivity and adaptability of LLMs. While the multi-agent setup introduces some computational complexity, as noted in \cite{feng2024modular}, we have addressed this by implementing optimization techniques such as efficient alignment \cite{dao2024flashattention, zhao2025swift} and inference \cite{kwon2023efficient}, effectively reducing computational burdens. As a result, our framework remains competitive with other cultural alignment techniques while ensuring better scalability.
Additionally, We envision ongoing efforts to distill the framework’s capabilities into smaller LLM variants, enabling more efficient deployment across a broader range of applications and devices. 

Finally, as highlighted by \cite{xu2025self,feng2024modular}, culture definition and cultural map division are inherently diverse and multifaceted. We validate our Cultural Palette with continent-to-country alignment based on continental cultural geography \cite{crang2013cultural}, offering a flexible framework that can be adapted to other cultural maps \cite{lechner2008world,inglehart2010wvs} and encouraging exploration of alternative cultural divisions to expand the applicability of our Cultural Palette \includegraphics[width=0.8\baselineskip]{fig/logo.png}.

\section*{Ethical Considerations}
This work builds on the public PRISM dataset~\cite{kirk2024prism} to construct the Pentachromatic Cultural Palette Dataset for cultural alignment, in full compliance with its usage policies. While aiming to promote inclusivity and reduce bias, we recognize potential risks of stereotyping or misrepresentation. To address this, all generated data undergo both model-in-the-loop and post-hoc human ethics reviews for cultural sensitivity following \citet{li2024culturellm}. Our framework advances pluralistic, equitable representation via multi-agent alignment and encourages continued expansion of culturally diverse perspectives in alignment research.

\bibliography{custom}

\begin{thebibliography}{54}
\providecommand{\natexlab}[1]{#1}

\bibitem[{Achiam et~al.(2023)Achiam, Adler, Agarwal, Ahmad, Akkaya, Aleman, Almeida, Altenschmidt, Altman, Anadkat et~al.}]{achiam2023gpt}
Josh Achiam, Steven Adler, Sandhini Agarwal, Lama Ahmad, Ilge Akkaya, Florencia~Leoni Aleman, Diogo Almeida, Janko Altenschmidt, Sam Altman, Shyamal Anadkat, et~al. 2023.
\newblock Gpt-4 technical report.
\newblock \emph{arXiv preprint arXiv:2303.08774}.

\bibitem[{Buschmann and Nolde(2018)}]{buschmann2018world}
Rainer~F Buschmann and Lance Nolde. 2018.
\newblock \emph{The World's Oceans: Geography, History, and Environment}.
\newblock Bloomsbury Publishing USA.

\bibitem[{Chan et~al.(2023)Chan, Garc{\'\i}a, Silvestri, O'Donnel, and Palla}]{chan2023harmonizing}
Alex~J Chan, Jos{\'e} Luis~Redondo Garc{\'\i}a, Fabrizio Silvestri, Colm O'Donnel, and Konstantina Palla. 2023.
\newblock Harmonizing global voices: Culturally-aware models for enhanced content moderation.
\newblock \emph{arXiv preprint arXiv:2312.02401}.

\bibitem[{Chen et~al.(2024)Chen, Chen, Rege, and Vinayak}]{chen2024pal}
Daiwei Chen, Yi~Chen, Aniket Rege, and Ramya~Korlakai Vinayak. 2024.
\newblock Pal: Pluralistic alignment framework for learning from heterogeneous preferences.
\newblock In \emph{NeurIPS 2024 Workshop on Fine-Tuning in Modern Machine Learning: Principles and Scalability}.

\bibitem[{Crang(2013)}]{crang2013cultural}
Mike Crang. 2013.
\newblock \emph{Cultural geography}.
\newblock Routledge.

\bibitem[{Dao(2024)}]{dao2024flashattention}
Tri Dao. 2024.
\newblock Flashattention-2: Faster attention with better parallelism and work partitioning.
\newblock In \emph{The Twelfth International Conference on Learning Representations}.

\bibitem[{Dubey et~al.(2024)Dubey, Jauhri, Pandey, Kadian, Al-Dahle, Letman, Mathur, Schelten, Yang, Fan et~al.}]{dubey2024llama}
Abhimanyu Dubey, Abhinav Jauhri, Abhinav Pandey, Abhishek Kadian, Ahmad Al-Dahle, Aiesha Letman, Akhil Mathur, Alan Schelten, Amy Yang, Angela Fan, et~al. 2024.
\newblock The llama 3 herd of models.
\newblock \emph{arXiv preprint arXiv:2407.21783}.

\bibitem[{DURMUS et~al.(2024)DURMUS, Nguyen, Liao, Schiefer, Askell, Bakhtin, Chen, Hatfield-Dodds, Hernandez, Joseph et~al.}]{durmus2024towards}
Esin DURMUS, Karina Nguyen, Thomas Liao, Nicholas Schiefer, Amanda Askell, Anton Bakhtin, Carol Chen, Zac Hatfield-Dodds, Danny Hernandez, Nicholas Joseph, et~al. 2024.
\newblock Towards measuring the representation of subjective global opinions in language models.
\newblock In \emph{First Conference on Language Modeling}.

\bibitem[{Feng et~al.(2024)Feng, Sorensen, Liu, Fisher, Park, Choi, and Tsvetkov}]{feng2024modular}
Shangbin Feng, Taylor Sorensen, Yuhan Liu, Jillian Fisher, Chan~Young Park, Yejin Choi, and Yulia Tsvetkov. 2024.
\newblock Modular pluralism: Pluralistic alignment via multi-llm collaboration.
\newblock In \emph{Proceedings of the 2024 Conference on Empirical Methods in Natural Language Processing}, pages 4151--4171.

\bibitem[{Goddard et~al.(2024)Goddard, Siriwardhana, Ehghaghi, Meyers, Karpukhin, Benedict, McQuade, and Solawetz}]{goddard2024arcee}
Charles Goddard, Shamane Siriwardhana, Malikeh Ehghaghi, Luke Meyers, Vladimir Karpukhin, Brian Benedict, Mark McQuade, and Jacob Solawetz. 2024.
\newblock Arcee’s mergekit: A toolkit for merging large language models.
\newblock In \emph{Proceedings of the 2024 Conference on Empirical Methods in Natural Language Processing: Industry Track}, pages 477--485.

\bibitem[{Guo et~al.(2025)Guo, Naous, Wakaki, Nishimura, Mitsufuji, Ritter, and Xu}]{guo2025care}
Geyang Guo, Tarek Naous, Hiromi Wakaki, Yukiko Nishimura, Yuki Mitsufuji, Alan Ritter, and Wei Xu. 2025.
\newblock Care: Aligning language models for regional cultural awareness.
\newblock \emph{arXiv preprint arXiv:2504.05154}.

\bibitem[{Haerpfer et~al.(2022)Haerpfer, Inglehart, Moreno, Welzel, Kizilova, Diez-Medrano, Lagos, Norris, Ponarin, and Puranen}]{wvs}
Christian Haerpfer, Ronald Inglehart, Alejandro Moreno, Christian Welzel, Kseniya Kizilova, Jaime Diez-Medrano, Marta Lagos, Pippa Norris, Eduard Ponarin, and Bjorn Puranen. 2022.
\newblock World values survey: Round seven-country-pooled datafile version 5.0.
\newblock \emph{Madrid, Spain \& Vienna, Austria: JD Systems Institute \& WVSA Secretariat}, 12(10):8.

\bibitem[{Han et~al.(2024)Han, Shenfeld, Srivastava, Kim, and Agrawal}]{han2024value}
Seungwook Han, Idan Shenfeld, Akash Srivastava, Yoon Kim, and Pulkit Agrawal. 2024.
\newblock Value augmented sampling for language model alignment and personalization.
\newblock \emph{arXiv preprint arXiv:2405.06639}.

\bibitem[{Hofstede(2009)}]{hofstede2009geert}
Geert Hofstede. 2009.
\newblock Geert hofstede cultural dimensions.

\bibitem[{Hong et~al.(2024)Hong, Lee, and Thorne}]{hong2024orpo}
Jiwoo Hong, Noah Lee, and James Thorne. 2024.
\newblock Orpo: Monolithic preference optimization without reference model.
\newblock In \emph{Proceedings of the 2024 Conference on Empirical Methods in Natural Language Processing}, pages 11170--11189.

\bibitem[{Hu et~al.(2022)Hu, Wallis, Allen-Zhu, Li, Wang, Wang, Chen et~al.}]{hu2022lora}
Edward~J Hu, Phillip Wallis, Zeyuan Allen-Zhu, Yuanzhi Li, Shean Wang, Lu~Wang, Weizhu Chen, et~al. 2022.
\newblock Lora: Low-rank adaptation of large language models.
\newblock In \emph{International Conference on Learning Representations}.

\bibitem[{Huang et~al.(2024)Huang, Yu, Zhu, Sun, Cheng, Dingjie, Chen, Alharthi, An, He et~al.}]{huang2024acegpt}
Huang Huang, Fei Yu, Jianqing Zhu, Xuening Sun, Hao Cheng, Song Dingjie, Zhihong Chen, Mosen Alharthi, Bang An, Juncai He, et~al. 2024.
\newblock Acegpt, localizing large language models in arabic.
\newblock In \emph{Proceedings of the 2024 Conference of the North American Chapter of the Association for Computational Linguistics: Human Language Technologies (Volume 1: Long Papers)}, pages 8132--8156.

\bibitem[{Ilharco et~al.(2023)Ilharco, Ribeiro, Wortsman, Schmidt, Hajishirzi, and Farhadi}]{ilharcoediting}
Gabriel Ilharco, Marco~Tulio Ribeiro, Mitchell Wortsman, Ludwig Schmidt, Hannaneh Hajishirzi, and Ali Farhadi. 2023.
\newblock Editing models with task arithmetic.
\newblock In \emph{The Eleventh International Conference on Learning Representations}.

\bibitem[{Inglehart and Welzel(2010)}]{inglehart2010wvs}
Ronald Inglehart and Chris Welzel. 2010.
\newblock The wvs cultural map of the world.
\newblock \emph{World Values Survey}.

\bibitem[{Insights(2021)}]{insights2021hofstede}
Hofstede Insights. 2021.
\newblock Hofstede cultural dimensions.

\bibitem[{Jang et~al.(2025)Jang, Yun, and Han}]{jang2025model}
Dong-Hwan Jang, Sangdoo Yun, and Dongyoon Han. 2025.
\newblock Model stock: All we need is just a few fine-tuned models.
\newblock In \emph{European Conference on Computer Vision}, pages 207--223. Springer.

\bibitem[{Jiang et~al.(2024)Jiang, Sablayrolles, Roux, Mensch, Savary, Bamford, Chaplot, Casas, Hanna, Bressand et~al.}]{jiang2024mixtral}
Albert~Q Jiang, Alexandre Sablayrolles, Antoine Roux, Arthur Mensch, Blanche Savary, Chris Bamford, Devendra~Singh Chaplot, Diego de~las Casas, Emma~Bou Hanna, Florian Bressand, et~al. 2024.
\newblock Mixtral of experts.
\newblock \emph{arXiv preprint arXiv:2401.04088}.

\bibitem[{Kirk et~al.(2024)Kirk, Whitefield, Rottger, Bean, Margatina, Mosquera-Gomez, Ciro, Bartolo, Williams, He et~al.}]{kirk2024prism}
Hannah~Rose Kirk, Alexander Whitefield, Paul Rottger, Andrew~M Bean, Katerina Margatina, Rafael Mosquera-Gomez, Juan Ciro, Max Bartolo, Adina Williams, He~He, et~al. 2024.
\newblock The prism alignment dataset: What participatory, representative and individualised human feedback reveals about the subjective and multicultural alignment of large language models.
\newblock \emph{Advances in Neural Information Processing Systems}, 37:105236--105344.

\bibitem[{Kojima et~al.(2022)Kojima, Gu, Reid, Matsuo, and Iwasawa}]{kojima2022large}
Takeshi Kojima, Shixiang~Shane Gu, Machel Reid, Yutaka Matsuo, and Yusuke Iwasawa. 2022.
\newblock Large language models are zero-shot reasoners.
\newblock \emph{Advances in neural information processing systems}, 35:22199--22213.

\bibitem[{Kwon et~al.(2023)Kwon, Li, Zhuang, Sheng, Zheng, Yu, Gonzalez, Zhang, and Stoica}]{kwon2023efficient}
Woosuk Kwon, Zhuohan Li, Siyuan Zhuang, Ying Sheng, Lianmin Zheng, Cody~Hao Yu, Joseph Gonzalez, Hao Zhang, and Ion Stoica. 2023.
\newblock Efficient memory management for large language model serving with pagedattention.
\newblock In \emph{Proceedings of the 29th Symposium on Operating Systems Principles}, pages 611--626.

\bibitem[{Lake et~al.(2024)Lake, Choi, and Durrett}]{lake2024distributional}
Thom Lake, Eunsol Choi, and Greg Durrett. 2024.
\newblock From distributional to overton pluralism: Investigating large language model alignment.
\newblock In \emph{Pluralistic Alignment Workshop at NeurIPS 2024}.

\bibitem[{Lawson and Qureshi(2024)}]{lawson2024merging}
Daniel Lawson and Ahmed~H Qureshi. 2024.
\newblock Merging decision transformers: Weight averaging for forming multi-task policies.
\newblock In \emph{2024 IEEE International Conference on Robotics and Automation (ICRA)}, pages 12942--12948. IEEE.

\bibitem[{Lechner and Boli(2008)}]{lechner2008world}
Frank~J Lechner and John Boli. 2008.
\newblock \emph{World culture: Origins and consequences}.
\newblock John Wiley \& Sons.

\bibitem[{Li et~al.(2024{\natexlab{a}})Li, Chen, Wang, Sitaram, and Xie}]{li2024culturellm}
Cheng Li, Mengzhuo Chen, Jindong Wang, Sunayana Sitaram, and Xing Xie. 2024{\natexlab{a}}.
\newblock Culturellm: Incorporating cultural differences into large language models.
\newblock \emph{Advances in Neural Information Processing Systems}, 37:84799--84838.

\bibitem[{Li et~al.(2024{\natexlab{b}})Li, Teney, Yang, Wen, Xie, and Wang}]{li2024culturepark}
Cheng Li, Damien Teney, Linyi Yang, Qingsong Wen, Xing Xie, and Jindong Wang. 2024{\natexlab{b}}.
\newblock Culturepark: Boosting cross-cultural understanding in large language models.
\newblock \emph{Advances in Neural Information Processing Systems}, 37:65183--65216.

\bibitem[{Li et~al.(2024{\natexlab{c}})Li, Wang, Zhang, and Zhao}]{li2024self}
Junlong Li, Jinyuan Wang, Zhuosheng Zhang, and Hai Zhao. 2024{\natexlab{c}}.
\newblock Self-prompting large language models for zero-shot open-domain qa.
\newblock In \emph{Proceedings of the 2024 Conference of the North American Chapter of the Association for Computational Linguistics: Human Language Technologies (Volume 1: Long Papers)}, pages 296--310.

\bibitem[{Li et~al.(2024{\natexlab{d}})Li, Zhou, Lipton, and Leqi}]{li2024personalized}
Xinyu Li, Ruiyang Zhou, Zachary~C Lipton, and Liu Leqi. 2024{\natexlab{d}}.
\newblock Personalized language modeling from personalized human feedback.
\newblock \emph{arXiv preprint arXiv:2402.05133}.

\bibitem[{Lin et~al.(2024)Lin, Ji, Tiedemann, Martins, and Sch{\"u}tze}]{lin2024mala}
Peiqin Lin, Shaoxiong Ji, J{\"o}rg Tiedemann, Andr{\'e}~FT Martins, and Hinrich Sch{\"u}tze. 2024.
\newblock Mala-500: Massive language adaptation of large language models.
\newblock \emph{arXiv preprint arXiv:2401.13303}.

\bibitem[{Madaan et~al.(2024)Madaan, Tandon, Gupta, Hallinan, Gao, Wiegreffe, Alon, Dziri, Prabhumoye, Yang et~al.}]{madaan2024self}
Aman Madaan, Niket Tandon, Prakhar Gupta, Skyler Hallinan, Luyu Gao, Sarah Wiegreffe, Uri Alon, Nouha Dziri, Shrimai Prabhumoye, Yiming Yang, et~al. 2024.
\newblock Self-refine: Iterative refinement with self-feedback.
\newblock \emph{Advances in Neural Information Processing Systems}, 36.

\bibitem[{Nicol and Macfarlane-Dick(2006)}]{nicol2006formative}
David~J Nicol and Debra Macfarlane-Dick. 2006.
\newblock Formative assessment and self-regulated learning: A model and seven principles of good feedback practice.
\newblock \emph{Studies in higher education}, 31(2):199--218.

\bibitem[{Ouyang et~al.(2022)Ouyang, Wu, Jiang, Almeida, Wainwright, Mishkin, Zhang, Agarwal, Slama, Ray et~al.}]{ouyang2022training}
Long Ouyang, Jeffrey Wu, Xu~Jiang, Diogo Almeida, Carroll Wainwright, Pamela Mishkin, Chong Zhang, Sandhini Agarwal, Katarina Slama, Alex Ray, et~al. 2022.
\newblock Training language models to follow instructions with human feedback.
\newblock \emph{Advances in neural information processing systems}, 35:27730--27744.

\bibitem[{Pourreza et~al.(2024)Pourreza, Sun, Li, Miculicich, Pfister, and Arik}]{pourreza2024sql}
Mohammadreza Pourreza, Ruoxi Sun, Hailong Li, Lesly Miculicich, Tomas Pfister, and Sercan~O Arik. 2024.
\newblock Sql-gen: Bridging the dialect gap for text-to-sql via synthetic data and model merging.
\newblock \emph{arXiv preprint arXiv:2408.12733}.

\bibitem[{Rafailov et~al.(2024)Rafailov, Sharma, Mitchell, Manning, Ermon, and Finn}]{rafailov2024direct}
Rafael Rafailov, Archit Sharma, Eric Mitchell, Christopher~D Manning, Stefano Ermon, and Chelsea Finn. 2024.
\newblock Direct preference optimization: Your language model is secretly a reward model.
\newblock \emph{Advances in Neural Information Processing Systems}, 36.

\bibitem[{Shetty et~al.(2025)Shetty, Beheshti, Dras, and Naseem}]{shetty-etal-2025-vital}
Anudeex Shetty, Amin Beheshti, Mark Dras, and Usman Naseem. 2025.
\newblock \href {https://aclanthology.org/2025.acl-long.1119/} {{VITAL}: A new dataset for benchmarking pluralistic alignment in healthcare}.
\newblock In \emph{Proceedings of the 63rd Annual Meeting of the Association for Computational Linguistics (Volume 1: Long Papers)}, pages 22954--22974, Vienna, Austria. Association for Computational Linguistics.

\bibitem[{Shi et~al.(2024)Shi, Li, Zhang, Ziems, Yu, Horesh, De~Paula, and Yang}]{shi2024culturebank}
Weiyan Shi, Ryan Li, Yutong Zhang, Caleb Ziems, Sunny Yu, Raya Horesh, Rog{\'e}rio~Abreu De~Paula, and Diyi Yang. 2024.
\newblock Culturebank: An online community-driven knowledge base towards culturally aware language technologies.
\newblock In \emph{Findings of the Association for Computational Linguistics: EMNLP 2024}, pages 4996--5025.

\bibitem[{Sileo(2024)}]{sileo-2024-tasksource}
Damien Sileo. 2024.
\newblock \href {https://aclanthology.org/2024.lrec-main.1361} {tasksource: A large collection of {NLP} tasks with a structured dataset preprocessing framework}.
\newblock In \emph{Proceedings of the 2024 Joint International Conference on Computational Linguistics, Language Resources and Evaluation (LREC-COLING 2024)}, pages 15655--15684, Torino, Italia. ELRA and ICCL.

\bibitem[{Sorensen et~al.(2024{\natexlab{a}})Sorensen, Jiang, Hwang, Levine, Pyatkin, West, Dziri, Lu, Rao, Bhagavatula et~al.}]{sorensen2024value}
Taylor Sorensen, Liwei Jiang, Jena~D Hwang, Sydney Levine, Valentina Pyatkin, Peter West, Nouha Dziri, Ximing Lu, Kavel Rao, Chandra Bhagavatula, et~al. 2024{\natexlab{a}}.
\newblock Value kaleidoscope: Engaging ai with pluralistic human values, rights, and duties.
\newblock In \emph{Proceedings of the AAAI Conference on Artificial Intelligence}, volume~38, pages 19937--19947.

\bibitem[{Sorensen et~al.(2024{\natexlab{b}})Sorensen, Moore, Fisher, Gordon, Mireshghallah, Rytting, Ye, Jiang, Lu, Dziri et~al.}]{sorensen2024position}
Taylor Sorensen, Jared Moore, Jillian Fisher, Mitchell Gordon, Niloofar Mireshghallah, Christopher~Michael Rytting, Andre Ye, Liwei Jiang, Ximing Lu, Nouha Dziri, et~al. 2024{\natexlab{b}}.
\newblock Position: a roadmap to pluralistic alignment.
\newblock In \emph{Proceedings of the 41st International Conference on Machine Learning}, pages 46280--46302.

\bibitem[{Tang et~al.(2024)Tang, Shen, Luo, Yin, Zhang, and Tao}]{tang2024merging}
Anke Tang, Li~Shen, Yong Luo, Nan Yin, Lefei Zhang, and Dacheng Tao. 2024.
\newblock Merging multi-task models via weight-ensembling mixture of experts.
\newblock In \emph{Forty-first International Conference on Machine Learning}.

\bibitem[{Vamplew et~al.(2024)Vamplew, Hayes, Foale, Dazeley, and Harland}]{vamplew2024multi}
Peter Vamplew, Conor~F Hayes, Cameron Foale, Richard Dazeley, and Hadassah Harland. 2024.
\newblock Multi-objective reinforcement learning: A tool for pluralistic alignment.
\newblock In \emph{Pluralistic Alignment Workshop at NeurIPS 2024}.

\bibitem[{Wang et~al.(2025)Wang, Jue, Athiwaratkun, Zhang, and Zou}]{wang2025mixture}
Junlin Wang, WANG Jue, Ben Athiwaratkun, Ce~Zhang, and James Zou. 2025.
\newblock Mixture-of-agents enhances large language model capabilities.
\newblock In \emph{The Thirteenth International Conference on Learning Representations}.

\bibitem[{Wang et~al.(2023)Wang, Kordi, Mishra, Liu, Smith, Khashabi, and Hajishirzi}]{wang2023self}
Yizhong Wang, Yeganeh Kordi, Swaroop Mishra, Alisa Liu, Noah~A Smith, Daniel Khashabi, and Hannaneh Hajishirzi. 2023.
\newblock Self-instruct: Aligning language models with self-generated instructions.
\newblock In \emph{Proceedings of the 61st Annual Meeting of the Association for Computational Linguistics (Volume 1: Long Papers)}, pages 13484--13508.

\bibitem[{Xu et~al.(2025)Xu, Leng, Yu, and Xiong}]{xu2025self}
Shaoyang Xu, Yongqi Leng, Linhao Yu, and Deyi Xiong. 2025.
\newblock Self-pluralising culture alignment for large language models.
\newblock In \emph{Proceedings of the 2025 Conference of the Nations of the Americas Chapter of the Association for Computational Linguistics: Human Language Technologies (Volume 1: Long Papers)}, pages 6859--6877.

\bibitem[{Yadav et~al.(2024)Yadav, Tam, Choshen, Raffel, and Bansal}]{yadav2024ties}
Prateek Yadav, Derek Tam, Leshem Choshen, Colin~A Raffel, and Mohit Bansal. 2024.
\newblock Ties-merging: Resolving interference when merging models.
\newblock \emph{Advances in Neural Information Processing Systems}, 36.

\bibitem[{Yang et~al.(2024{\natexlab{a}})Yang, Yang, Hui, Zheng, Yu, Zhou, Li, Li, Liu, Huang et~al.}]{yang2024qwen2}
An~Yang, Baosong Yang, Binyuan Hui, Bo~Zheng, Bowen Yu, Chang Zhou, Chengpeng Li, Chengyuan Li, Dayiheng Liu, Fei Huang, et~al. 2024{\natexlab{a}}.
\newblock Qwen2 technical report.
\newblock \emph{arXiv preprint arXiv:2407.10671}.

\bibitem[{Yang et~al.(2024{\natexlab{b}})Yang, Shen, Guo, Wang, Cao, Zhang, and Tao}]{yang2024model}
Enneng Yang, Li~Shen, Guibing Guo, Xingwei Wang, Xiaochun Cao, Jie Zhang, and Dacheng Tao. 2024{\natexlab{b}}.
\newblock Model merging in llms, mllms, and beyond: Methods, theories, applications and opportunities.
\newblock \emph{arXiv preprint arXiv:2408.07666}.

\bibitem[{Zhao et~al.(2025)Zhao, Huang, Hu, Wang, Mao, Zhang, Jiang, Wu, Ai, Wang et~al.}]{zhao2025swift}
Yuze Zhao, Jintao Huang, Jinghan Hu, Xingjun Wang, Yunlin Mao, Daoze Zhang, Zeyinzi Jiang, Zhikai Wu, Baole Ai, Ang Wang, et~al. 2025.
\newblock Swift: a scalable lightweight infrastructure for fine-tuning.
\newblock In \emph{Proceedings of the AAAI Conference on Artificial Intelligence}, volume~39, pages 29733--29735.

\bibitem[{Zheng et~al.(2023)Zheng, Zhou, Meng, Zhou, and Huang}]{zheng2023large}
Chujie Zheng, Hao Zhou, Fandong Meng, Jie Zhou, and Minlie Huang. 2023.
\newblock Large language models are not robust multiple choice selectors.
\newblock In \emph{The Twelfth International Conference on Learning Representations}.

\bibitem[{Zhou et~al.(2023)Zhou, Zhu, Chen, Chen, Zhao, Chen, Lin, Wen, and Han}]{zhou2023don}
Kun Zhou, Yutao Zhu, Zhipeng Chen, Wentong Chen, Wayne~Xin Zhao, Xu~Chen, Yankai Lin, Ji-Rong Wen, and Jiawei Han. 2023.
\newblock Don't make your llm an evaluation benchmark cheater.
\newblock \emph{arXiv preprint arXiv:2311.01964}.

\end{thebibliography}
\newpage
\appendix

\section*{Appendix of Cultural Palette}
This supplementary material provides supplementary materials: Appendix~A describes the construction details of Pentachromatic Cultural Palette dataset, Appendix~B details the experimental setup and hyperparameters, Appendix~C presents qualitative case studies on cultural alignment across \textit{w/ NLI evaluation} (\ref{subsec:culturealign}) and \textit{w/ GlobalOpinionQA} (\ref{subsec:qa}), and Appendix~D provides an additional quantitative continent-to-country correlation analysis with Modular Pluralism \cite{feng2024modular}, further validating Cultural Palette’s effectiveness in bridging continent-level reasoning with country-level cultural alignment.

\section{Details for Pentachromatic Cultural Palette Dataset}
\subsection{Prompt for Pentachromatic Cultural Palette Dataset}
\label{subsec:dataprompt}
\begin{center}
\begin{tcolorbox}[width=\linewidth,colback=white, title=Step1: Continent-aware Responses]
\label{prompt}
{\slshape 
\textbf{Instruction}\\
You are a knowledgeable chatbot specializing in \{Continent\}. Provide concise, accurate, and culturally sensitive responses tailored to user queries about \{Continent\}'s geography, culture, history, and current affairs. Ensure your responses demonstrate respect and understanding of the diverse cultures within \{Continent\}. \\
{\textbf{1. Understand Query Context:}}\\ 
Identify the topic and cultural relevance of the user's query (e.g., geography, culture, history, or current events).\\
{\textbf{2. Generate Continent-aware Response:}}\\ 
Show respect for the diverse cultures, avoiding assumptions or biased statements. Highlight the uniqueness and importance of specific cultural aspects when appropriate.\\
\textbf{Examples:} \\
Question: \{Question\}  Answer: \{Answer\} \\
\textbf{Input:} \{Question\} \\
\textbf{Output:} \{Continent-aware Response\}
}
\end{tcolorbox}
\noindent\begin{minipage}{\linewidth}
\captionof{figure}{Prompt for continent-aware responses}\label{promptstep1}
\end{minipage}
\end{center}

\begin{center}
\begin{tcolorbox}[width=\linewidth,colback=white, title=Step2: Cross-continent Feedback]
\label{prompt}
{\slshape 
\textbf{Instruction}\\
Review the response for cultural clarity, ensuring \{Continent\} perspectives are prominent and free from interference by other cultures \{Other Continents\}. If confusion arises, provide improvement suggestions using Hofstede's cultural dimensions: \\
{\textbf{1. Analyze Query Context:}}\\ 
Assess whether the response fully aligns with the cultural, historical, and geographical relevance of \{Continent\}. Identify any unnecessary elements or overlaps with other continents' cultural markers.\\
{\textbf{2. Refine Response for \{Continent\} Perspective:}}\\ 
Emphasize key cultural, historical, or environmental aspects unique to \{Continent\}. Remove any elements that might introduce cultural interference or bias. Ensure the response highlights community, tradition, and respect for local customs. \\
{\textbf{3. Apply Hofstede's Cultural Dimensions Theory for Feedback:}}\\ 
- \textbf{Power Distance (PDI):} Does the response reflect the balance between leadership (e.g., village chiefs) and community roles? \\
- \textbf{Individualism vs. Collectivism (IDV):} Does it highlight collective well-being and community-driven decisions? \\
- \textbf{Masculinity vs. Femininity (MAS):} Does it prioritize care, cooperation, and harmony over competition? \\
- \textbf{Uncertainty Avoidance (UAI):} Does it address Oceania's adaptability to change while respecting traditions? \\
- \textbf{Long-Term vs. Short-Term Orientation (LTO):} Does it balance preservation of traditions with immediate adaptability? \\
- \textbf{Indulgence vs. Restraint (IVR):} Does it reflect the expressive and celebratory aspects of Oceanian cultures? \\
\textbf{Input:} \{Continent-aware Responses\} \\
\textbf{Output:} \{Cross-continent Feedback\}
} 
\end{tcolorbox}
\noindent\begin{minipage}{\linewidth}
\captionof{figure}{Prompt for cross-continent feedback}\label{promptstep2}
\end{minipage}
\end{center}
\begin{center}
\begin{tcolorbox}[width=\linewidth,colback=white, title=Step3: Continent-specific Aggregation]
\label{promptstep3}
{\slshape 
\textbf{Instruction}\\
You are tasked with improving the following response for the query \{query\} by incorporating the cross-continent review feedback. \\
{\textbf{1. Review Context Response:}}\\ 
Carefully read and analyze the Base Response and Review Feedback provided. Identify elements in the base response that lack focus on \{Continent\} or introduce cultural elements from other continents.\\
{\textbf{2. Apply Review Feedback:}}\\ 
Incorporate all suggestions from the review feedback into the base response.
Ensure the revised response highlights \{Continent\}-specific culture, history, and values and focus on \{Continent\} Perspectives, avoiding any overlap with other cultural contexts.\\
\textbf{Input:} \\
Base Response: \{Base response\} \\
Review Feedback: \{Review feedback\} \\
\textbf{Output:} \{Continent\-focused Response\}
} 
\end{tcolorbox}
\noindent\begin{minipage}{\linewidth}
\captionof{figure}{Prompt for Continent-specific Aggregation}\label{promptstep3}
\end{minipage}
\end{center}

\begin{center}
\begin{tcolorbox}[width=\linewidth,colback=white, title=Step4: Self-Judge Refinement]
\label{promptstep4}
{\slshape 
\textbf{Instruction}\\
You are now a cultural self-judge. Your task is to assess whether the aggregated response reflects accurate, respectful, and continent-specific perspectives for the user query. Critically evaluate the revised response based on feedback, and decide whether further improvement is needed.\\
\textbf{1. Evaluate Cultural Fidelity:}\\
Check if the revised response aligns with the core values, traditions, and worldview of \{Continent\}. Is the response focused, respectful, and specific? Does it avoid generalizations or unrelated cross-cultural references?

\textbf{2. Verify Feedback Integration:}\\
Determine whether the suggestions from cross-continent feedback have been fully addressed. Are the earlier critiques resolved in this version?

\textbf{3. Quality Criteria for Acceptance:}\\
Approve the response only if it meets all the following:  
- Emphasizes cultural clarity and salience  
- Integrates prior feedback  
- Avoids cultural interference  
- Is concise, coherent, and informative

\textbf{Output:}\\
If the response is acceptable, return:  
\textit{[Approved]}  

If the response requires revision, return:  
\textit{[Revise]}  
And explain what needs to be changed.
}
\end{tcolorbox}
\noindent\begin{minipage}{\linewidth}
\captionof{figure}{Prompt for self-judge refinement to ensure cultural alignment}\label{promptstep4}
\end{minipage}
\end{center}

\subsection{Distribution in Pentachromatic Cultural Palette Dataset}
\label{subsec:distribution}
\begin{figure*}[t]
  \centering
  \includegraphics[width=\linewidth]{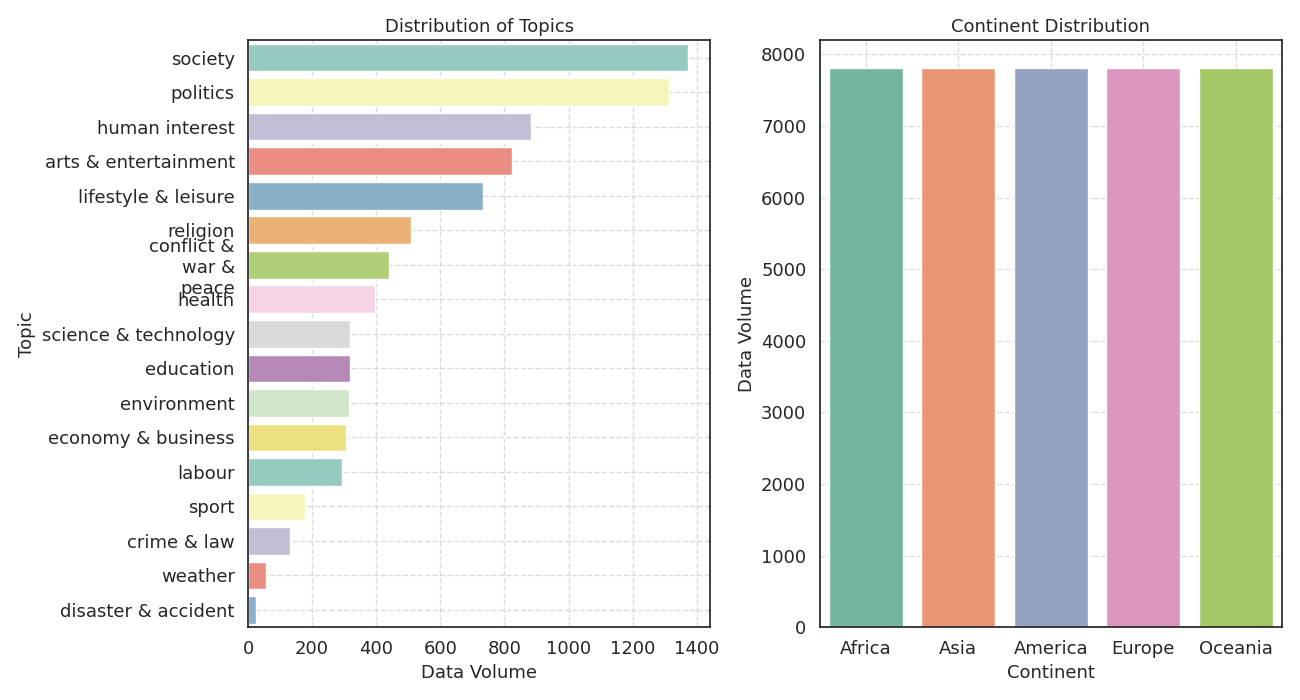}
  \caption{Distribution of topics and continents in Pentachromatic Cultural Palette Dataset.}
  \label{fig:distribution}
\end{figure*}

Figure~\ref{fig:distribution} illustrates the distribution of topics and continent over our Pentachromatic Cultural Palette Dataset. The topics were categorized using a topic modeling llm \footnote{https://huggingface.co/classla/multilingual-IPTC-news-topic-classifier, License:
cc-by-sa-4.0} which classifies the questions into distinct thematic clusters. The distribution of continents, on the other hand, is evenly balanced, with an equal number of questions synthesized for each continent: Africa, America, Asia, Europe, and Oceania.

\section{Experiment Details}
\label{sec:experiment}
\paragraph{Experiment Setting.} We adopt the ORPO method \cite{hong2024orpo} with default hyperparameters for Llama3.1-8B-Instruct \cite{dubey2024llama} and Qwen2.5-7B-Instruct \cite{yang2024qwen2}, using a learning rate of 5e-5 for 2 epochs. 
We select the average result from five runs as the experimental results.
\begin{table*}[!ht]
    \centering
    \begin{tabular}{llll}
    \toprule
    Resources & Type & Link & License \\
    \midrule
    MergeKit \cite{goddard2024arcee} & Code & {https://github.com/arcee-ai/mergekit} & LGPL-3.0 license \\
    MS-Swift \cite{zhao2025swift} & Code & {https://github.com/modelscope/ms-swift} & Apache-2.0 license \\
    \bottomrule
    \end{tabular}
    \captionsetup{width=\textwidth}
    \caption{Licenses for the software libraries utilized in our  work.}
    \label{tab:licenses}
\end{table*}
\paragraph{Alignment Implement.} We align continent-specific LLMs using ORPO \cite{hong2024orpo}, optimizing the following loss, $\mathcal{L}_{\text{ORPO}}$:

\begin{small}
\begin{equation}
\label{eq:orpo_loss_merged}
\begin{aligned}
\boldsymbol{\mathcal{L}_{\text{ORPO}}} 
&= \frac{1}{n} \sum_{i=1}^{n} \Biggl[
\underbrace{\boldsymbol{\ell}(\text{Pref}^{(i)}, \text{Query}^{(i)})} \\
&\quad + \boldsymbol{\lambda} \underbrace{\sum_{k=1}^{4} \left( -\log \sigma \left( \log \frac{\boldsymbol{p_\theta}(\text{Pref}^{(i)} \mid \text{Query}^{(i)})}{\boldsymbol{p_\theta}(\text{Rej}_k^{(i)} \mid \text{Query}^{(i)})} \right) \right)}_{\text{(b) Cross-continent Contrastive Alignment}} \Biggr] 
\end{aligned}
\end{equation}
\end{small}

where $\ell(\text{Pref}^{(i)}, \text{Query}^{(i)})$ denotes the continent-specific supervised fine-tuning (SFT) loss, aligning the LLM to generate the preferred response $\text{Pref}^{(i)}$ given query $\text{Query}^{(i)}$. The second term enforces cross-continent contrastive alignment by penalizing less-preferred responses $\text{Rej}_k^{(i)}$ from other continents that diverge from the preference of the target region. We set the odds ratio $\lambda=0.1$, following MS-Swift \cite{zhao2025swift}, to balance in-continent preference alignment with intercontinental contrastive separation.

\paragraph{Licenses.}The licenses for the codes used in our work are listed in Table~\ref{tab:licenses}.

\newpage
\section{Case Study}

\subsection{Case Study for Cultural Alignment w/ \textit{NLI evaluation}}\label{subsec:culturealign}
\label{subsec:nli}
To validate our Cultural Palette's efficacy, we conducted a case study (Tables~\ref{tab:happen1} \& \ref{tab:happen2}) addressing the question, "What happens when we die?" , to demonstrate how our Cultural Palette generates diverse responses across continents, reflecting each region's unique perspective on death. In Oceania, the reunion of souls with ancestors is emphasized, highlighting the cultural significance of these traditions. In Africa, the continuity of the soul and its connection to ancestors reinforce communal bonds. In Asia, beliefs such as reincarnation, enlightenment, and ancestor worship illustrate varied views of the afterlife. In America, a blend of perspectives is presented, including concepts of the afterlife and views on the cessation of consciousness. Finally, Europe encompasses various death-related traditions, showcasing the diversity of beliefs about life after death. This analysis underscores the model’s ability to capture cultural contexts and reveals distinct views on death across continents.

\subsection{Case Study for Cultural Palette \textit{w/ GlobalOpinionQA}}
\label{subsec:qa}
We provide a case study for Cultural Palette \textit{w/ GlobalOpinionQA} to demonstrate how our Cultural Palette works, as detailed in Table~\ref{tab:household_finances}.

\begin{table*}[ht]
    \centering
    \fontsize{11}{1}\selectfont
    \renewcommand{\arraystretch}{1}
    \begin{tabularx}{\textwidth}{p{0.08\textwidth}|p{0.39\textwidth}|p{0.35\textwidth}|p{0.09\textwidth}}
        \toprule
        \textbf{Continent} & \textbf{Input: What happens when we die?} & \textbf{Ground Truth} &  \textbf{$S_{semantic}$} \\
        \midrule
        \textbf{Oceania} & In Oceania, the concept of death and the afterlife varies among different cultures and communities. \highlightOceania{\textbf{Many indigenous cultures in Oceania believe in a spiritual realm where the deceased continue to exist and play an active role in the lives of the living.}} For example, in some Pacific Island cultures, \highlightOceania{\textbf{the spirits of ancestors are believed to guide and protect their descendants, while in other cultures, the deceased are thought to journey to a spiritual realm where they are judged based on their actions in life.}} & In Oceania, the beliefs surrounding what occurs after death are rich and diverse. Among the indigenous communities, \highlightOceania{\textbf{there is a profound connection to an afterlife where spirits are reunited with their ancestors in a sacred land}}. Additionally, some cultures embrace the idea of \highlightOceania{\textbf{reincarnation or embarking on a spiritual voyage to another realm beyond our earthly existence.}} It is paramount to approach these beliefs with deep respect and understanding, honoring the profound significance they hold within \highlightOceania\textbf{{Oceanian cultures}}. & 0.8927 \\

        \midrule
        \textbf{Africa} & In many African cultures, \highlightAfrica{\textbf{the concept of death is deeply intertwined with spirituality and the afterlife.}} The belief in an afterlife varies among different African societies, but it is often seen as \highlightAfrica{\textbf{a continuation of life in a spiritual realm.}} The deceased are believed to journey to the spirit world, where they are judged based on their deeds in life. \highlightAfrica{\textbf{The spirits of the ancestors are revered and believed to play a significant role in guiding the living}.} & In various African cultures, there is a profound belief in the continuity of life beyond death. \highlightAfrica{\textbf{Ancestral spirits play a significant role, where the departed are thought to watch over and guide their living relatives.} }This connection with ancestors is cherished, and rituals are performed to honor and communicate with them. Additionally, some traditions believe in \highlightAfrica{\textbf{reincarnation, suggesting that the soul returns in a new form after death.}} The respect for these diverse beliefs reflects the communal values deeply rooted in many \highlightAfrica{\textbf{African societies}}, honoring the interconnectedness between the living and the deceased. &  0.7671 \\

        \midrule
        \textbf{Asia} & In many Asian cultures, the concept of death is deeply \highlightAsia{\textbf{intertwined with spiritual beliefs and traditions.}} The idea of what happens after death varies across different cultures and religions in Asia. For example, in Buddhism, the belief is that \highlightAsia{\textbf{the soul is reborn into a new existence based on past actions, known as karma.}} In Hinduism, the soul is believed to be \highlightAsia{\textbf{reincarnated into a new life cycle, influenced by the law of karma.}} In many Asian cultures, the deceased are often believed to \highlightAsia{\textbf{continue their journey in the afterlife, where they may be reunited with ancestors or other loved ones.}} It's important to note that these beliefs can vary widely across different cultures and individuals in Asia. & In Asian cultures, diverse beliefs shape perspectives on death and the afterlife. Hinduism teaches that \highlightAsia{\textbf{souls undergo reincarnation driven by karma, seeking spiritual growth}}. Buddhism emphasizes \highlightAsia{\textbf{breaking free from the cycle of rebirth through enlightenment.}} In Japan, \highlightAsia{\textbf{ancestor veneration and a belief in spirits existing in the afterlife are prominent practices.}} The rich tapestry of Asian beliefs offers varied insights into what lies beyond our earthly existence. &  0.5847 \\
        \bottomrule
    \end{tabularx}
    \caption{Case Study for Cultural Platette on Cultural Alignment w/ \textit{NLI evaluation} (A).}
    \label{tab:happen1}
\end{table*}
\clearpage
\begin{table*}[ht]
    \centering
    \fontsize{11}{1}\selectfont
    \renewcommand{\arraystretch}{1}
    \begin{tabularx}{\textwidth}{p{0.08\textwidth}|p{0.34\textwidth}|p{0.32\textwidth}|p{0.16\textwidth}}
        \toprule
        \textbf{Continent} & \textbf{Input: What happens when we die?} & \textbf{Ground Truth} &   \textbf{$S_{semantic}$} \\
        \midrule
        \textbf{America} & In American culture, the concept of what happens after death is \highlightAmerica{\textbf{deeply rooted in various beliefs and traditions.}} Many Americans find \highlightAmerica{\textbf{solace in the idea of an afterlife, where the soul continues to exist in a spiritual realm.}} This belief is often \highlightAmerica{\textbf{tied to religious teachings, such as Christianity, Judaism, Islam, and others.}} Some believe in \highlightAmerica{\textbf{reincarnation}}, while others may \highlightAmerica{\textbf{view death as the end of physical existence.}} It's important to acknowledge that these perspectives vary widely among individuals and communities in America. & In American culture, beliefs about what happens after death can vary widely. Some Americans \highlightAmerica{\textbf{adhere to religious beliefs that include an afterlife in heaven or hell}}, while others believe in concepts like \highlightAmerica{\textbf{reincarnation}}. Additionally, some Americans \highlightAmerica{\textbf{view death from a more secular perspective where consciousness simply ceases to exist.}} & 0.7557 \\
        \midrule
        \textbf{Europe} & In Europe, the concept of what happens after death is deeply rooted in various cultural, religious, and philosophical beliefs. \highlightEurope{\textbf{The idea of an afterlife, where the soul or spirit continues to exist beyond physical death, is a common thread in many European cultures.}} For example, \highlightEurope{\textbf{in Christianity, the belief in heaven and hell is prevalent}}, while in Islam, \highlightEurope{\textbf{the concept of Jannah (paradise) and Jahannam (hell) is central.}} Additionally, some European cultures \highlightEurope{\textbf{believe in reincarnation or the transmigration of the soul.}} It's important to note that these beliefs can vary significantly across different European countries and communities. & In Europe, the beliefs regarding what happens when we die are deeply rooted in diverse cultural and religious traditions. For instance, \highlightEurope{\textbf{Christian belief in heaven and hell has been significant throughout European history, shaping perspectives on the afterlife.}} Similarly, \highlightEurope{\textbf{Nordic cultures have a rich tradition of Valhalla, a majestic hall where warriors go after death.}} Furthermore, the \highlightEurope{\textbf{Greek belief in the underworld, ruled by gods like Hades and Persephone, adds another layer to European perceptions of death.}} These varied beliefs reflect the intricate tapestry of European culture and highlight the importance of understanding and respecting different perspectives on life's end. & 0.7085
        \\
        \bottomrule
    \end{tabularx}
    \caption{Case Study for Cultural Platette on Cultural Alignment w/ \textit{NLI evaluation} (B).}
    \label{tab:happen2}
\end{table*}

\clearpage

\begin{table}[H]
    \fontsize{11}{1}\selectfont
    \renewcommand{\arraystretch}{1.05}
    \begin{tabularx}{\textwidth}{l|X}
        \toprule
        \textbf{Cultural Palette} & \textbf{Step} \\
        \midrule
        \textbf{Question} & Question: Most people consider both freedom and equality to be important, but if you had to choose between them, which one would you consider more important?
        \newline
        Options: \textbf{A}. Freedom \textbf{B}. Equality \textbf{C}. Don't know \textbf{D}. No answer\\
        \midrule
        \textbf{Draft} & 
        \highlightAfrica{\textbf{African Culture perspective}}: From an African perspective, the concept of equality is deeply rooted in many African cultures, where community and collective well-being are highly valued. In this context, prioritizing equality over freedom is often seen as a way to promote social cohesion, reduce inequality, and ensure that everyone has a fair chance to succeed ... This philosophy suggests that individual freedom is not as important as the collective well-being of the community. 
        \textbf{B. Equality}
        \newline
        
        \highlightAmerica{\textbf{America Culture perspective}}: From an American perspective, considering New Zealand's unique culture, traditions, and history, one might argue that equality is more important. ... Their focus on freedom reflects a commitment to individual rights and autonomy, which are core values in New Zealand's society. While equality is undoubtedly crucial, the context of New Zealand's history and culture emphasizes the importance of freedom in fostering a just and inclusive society. 
        \newline
        \textbf{A. Freedom}
        \newline
        
        \highlightOceania{\textbf{Oceania Culture perspective}}: From an Oceania perspective, New Zealand's unique culture, traditions, and history emphasize the importance of freedom. ... In the Pacific region, the concept of "manaakitanga" (respect and care for others) is deeply rooted in many cultures, but it is often balanced with the importance of individual freedom and autonomy. New Zealand's emphasis on freedom reflects this balance, prioritizing the rights and freedoms of individuals while also promoting social cohesion and community. 
        \textbf{A. Freedom}
        \newline
        
        \highlightEurope{\textbf{Europe Culture perspective}}: From a European perspective, New Zealand's unique culture, traditions, and history highlight a strong emphasis on both freedom and equality, reflecting its Māori and European influences. ... However, given the options provided, the most appropriate answer would be \textbf{C. Don't know}, as the question presents a false dichotomy. In New Zealand's context, both freedom and equality are considered paramount, and the country's history and culture do not necessitate a choice between them. 
        \textbf{C. Don't know}
        \newline
        
        \highlightAsia{\textbf{Asia Culture perspective}}: From an Asian perspective, New Zealand's emphasis on equality is particularly relevant in its multicultural society ... In this context, equality is considered more important than freedom, as it ensures that all individuals have equal opportunities and are treated with dignity and respect. This aligns with the values of many Asian cultures, which prioritize harmony and social cohesion over individual freedom. 
        \textbf{B. Equality}

        \\
        \midrule
        \textbf{Regulate} & 
        Considering the diverse cultural perspectives, New Zealand's cultural attributes can be summarized as community and collective well-being, freedom, and multiculturalism... Many cultures, including African, Asian, and Pacific Islander, prioritize community and collective well-being, while others, such as American and European, value individual freedom and autonomy... the Oceania perspective emphasizes the importance of freedom... Individual freedom is essential to promoting social cohesion and community, and New Zealand's emphasis on freedom reflects a commitment to individual rights and autonomy. ...
        \textbf{A.} Freedom
        \\
        \midrule
        \textbf{Answer} & 
        \textbf{A.} Freedom \textcolor{green}{\checkmark}
        \newline
        \textbf{Gold Distribution:} [0.6709999999999999, 0.242, 0.061, 0.0, 0.026000000000000002]
        \newline
        \textbf{Pred Distribution:} [0.6049056212210604, 0.3230594648058141, 0.060349139731253465, 0.0116857742428727]
        \\
        \bottomrule
    \end{tabularx}
    
    \begin{fullpagecaption}{\label{tab:household_finances}Cultural Alignment on Decision-Making in Household Finances}
    \end{fullpagecaption}

\end{table}

\clearpage
\section{Comparative Analysis via Pearson Correlation Across Countries}
\label{subsec:pearson}
\begin{figure}[h]
  \centering
  \includegraphics[width=\linewidth]{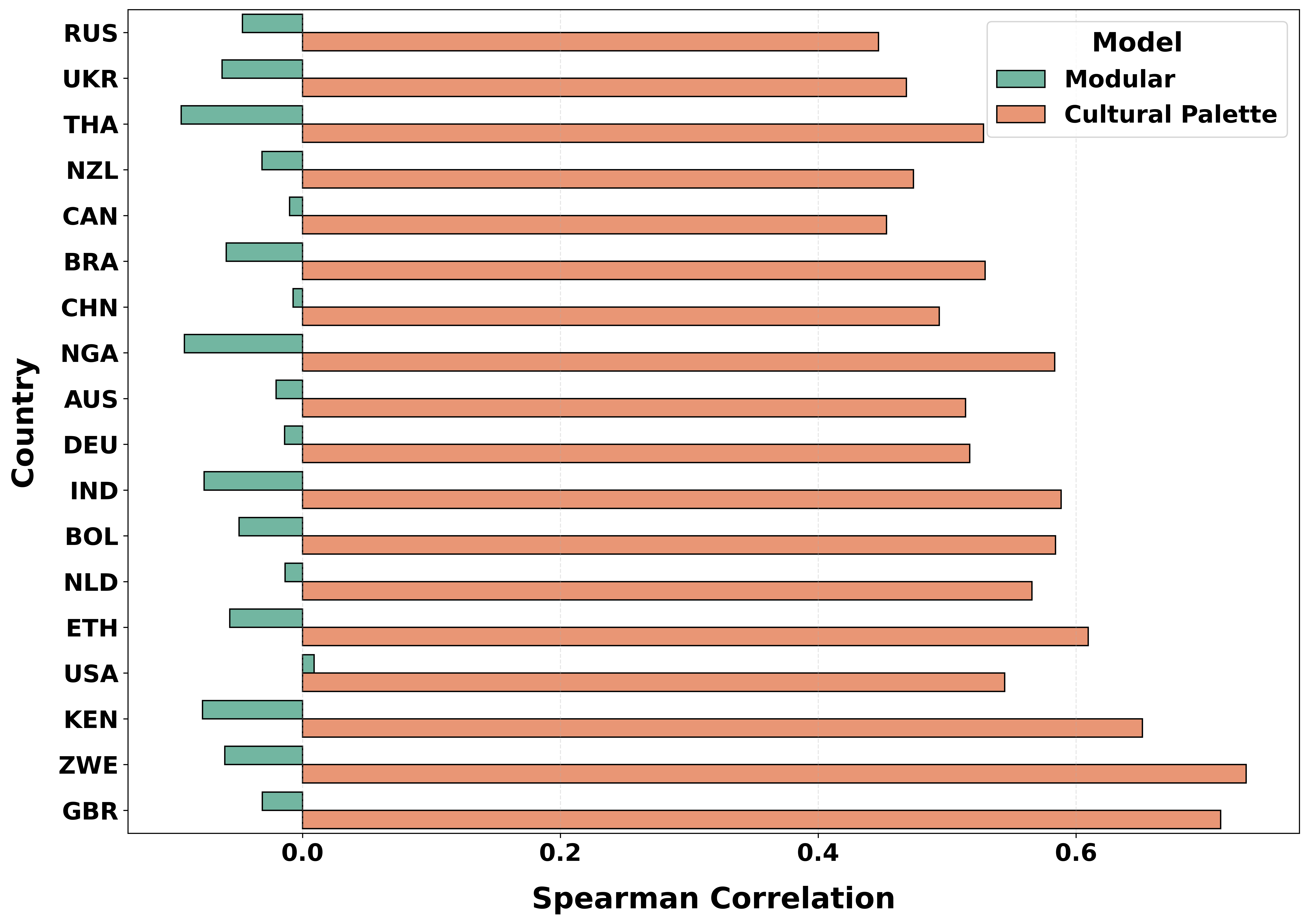}
  \caption{Comparative Analysis of Modular Pluralism \cite{feng2024modular} and Cultural Palette Models via Pearson Correlation Across Countries}
  \label{fig:comp}
\end{figure}
The comparison detailed in Fig.~\ref{fig:comp} illustrates that the Cultural Palette model consistently captures cultural and regional nuances more effectively across diverse countries such as Ethiopia, India, and Kenya, where it achieves notably higher Pearson correlations. This suggests strong adaptability to local contexts and complex sociocultural patterns. In contrast, Modular Pluralism \cite{feng2024modular} struggles in these countries, often producing low or near-zero correlations, highlighting its limited ability to model cultural heterogeneity within and across nations. Additionally, even in countries like Russia, China, and Ukraine, the Cultural Palette maintains a clear advantage, reinforcing the importance of culturally aware modeling. These qualitative insights emphasize that incorporating country-specific cultural factors is critical for building robust predictive models in global, multicultural environments.
\end{document}